\documentclass[journal]{IEEEtran}
\ifCLASSINFOpdf
\else
\fi
\usepackage{amsmath}
\usepackage{amsthm}
\newtheorem{theorem}{Theorem}
\newtheorem{lemma}{Lemma}
\newtheorem{definition}{Definition}
\newtheorem{prop}{Proposition}
\newtheorem{manualtheoreminner}{Theorem}

\usepackage{algorithm}
\usepackage{algorithmic}
\usepackage{microtype}
\usepackage{graphicx}
\usepackage{subfigure}
\usepackage{booktabs} 
\usepackage{amsfonts}
\usepackage{xcolor}
\usepackage{nccmath}
\usepackage{bm}
\usepackage{bbm}
\usepackage{xr}
\usepackage{array}
\usepackage{multirow}
\newcolumntype{P}[1]{>{\centering\arraybackslash}p{#1}}
\DeclareMathOperator*{\argmin}{arg\,min}

\begin{document}
%
\title{Multi-Agent Trust Region Policy Optimization}
%
%
%

\author{Hepeng~Li,~\IEEEmembership{Student Member,~IEEE,}
        and~Haibo~He,~\IEEEmembership{Fellow,~IEEE,}
\thanks{This material is based upon work supported by the National Science Foundation under Grant No. ECCS 1917275. (Corresponding author: Haibo He.)}
\thanks{Hepeng Li and Haibo He were with the Department of Electrical, Computer and Biomedical Engineering, University of Rhode Island, South Kingstown, RI, USA e-mail: (hepengli@uri.edu, haibohe@uri.edu).}
}

%
%

\markboth{}
{Shell \MakeLowercase{\textit{et al.}}: Bare Demo of IEEEtran.cls for IEEE Journals}
%



\maketitle

\begin{abstract}
We extend trust region policy optimization (TRPO) \cite{pmlr-v37-schulman15} to cooperative multi-agent reinforcement learning (MARL) for partially observable Markov games (POMG). We show that the policy update rule in TRPO can be equivalently transformed into a distributed consensus optimization for networked agents when the agents' observation is sufficient. By using local convexification and trust-region method, we propose a fully decentralized MARL algorithm based on a distributed alternating direction method of multipliers (ADMM). During training, agents only share local policy  ratio with neighbors via a peer-to-peer communication network. Compared with traditional centralized training methods in MARL, the proposed algorithm does not need a control center to collect global information, such as global state, collective reward, or shared policy and value network parameters. Experiments on two cooperative environments demonstrate the effectiveness of the proposed method.
\end{abstract}


%
\IEEEpeerreviewmaketitle

\section{Introduction}
%
%
%
%
\IEEEPARstart{R}{einforcement} learning (RL) holds promise in achieving artificial general intelligence (AGI) \cite{SILVER2021103535} as deep learning (DL) technologies \cite{lecun2015deep} help unleash its power from the curse of dimensionality. The combination of RL and DL promotes the development of deep RL (DRL) approaches, e.g. deep Q-network (DQN) \cite{13125602}, deep deterministic policy gradient (DDPG) \cite{pmlr-v32-silver14}, proximal policy optimization (PPO) \cite{170706347}, TRPO \cite{pmlr-v37-schulman15}, asynchronous advantage actor-critic (A3C) \cite{160201783}, soft actor-critic (SAC) \cite{pmlr-v80-haarnoja18b}. These approaches enable us to create agents that can learn and adapt to new tasks and environments similarly to humans. DRL algorithms have been successfully applied in a wide range of challenging decision-making tasks, such as playing games \cite{SilverHuangEtAl16nature}, controlling robots \cite{2946684}, and predicting protein structure \cite{AlphaFold2021}.

Recent development in multi-agent RL (MARL) has facilitated the collaboration of a group of agents by maximizing a collective reward. This extends the traditional RL paradigm from single-agent domains to multi-agent systems (MAS) and is useful in a wide range of applications, such as multiplayer games \cite{samvelyan19smac}, coordinating self-driving vehicles \cite{li2018reinforcement}, multi-robot systems \cite{Laetitia2012}, traffic signal control \cite{Wang2020}, and smart grids \cite{Manisa2009}.

Most existing MARL algorithms can be classified into three categories: (1) \emph{fully centralized}, such as joint action learning \cite{Claus1998}, which learns a centralized policy by reducing the problem to a single-agent RL over the global observation and action space, (2) \emph{centralized training decentralized execution}, such as the approach proposed by Lowe et al. \cite{lowe2020multiagent}, which learns local policies (that select actions based on local observations) in a centralized manner by using global information, such as aggregated observations, joint rewards, policy parameters or gradients of other agents, etc., and (3) \emph{fully decentralized}, such as independent learning \cite{Ming1997}, which learns local policies in a decentralized manner, and does not require global information in the training and execution stages.

\emph{Centralized training decentralized execution} is preferred in many studies since the learned policies can be executed in a distributed way while addressing the \emph{nonstationarity} issue \cite{Matignon2012} that arises when agents learn policies independently. \emph{Nonstationarity} issue arises when the agents learn policies independently. In the independent learning setting, if one agent changes its policy, the state transition and reward perceived by others will change since the agent is part of the environment from the perspective of others. If all agents change policy independently, the environment perceived by every agent would be non-stationary, making it difficult to converge. Centralized training addresses this issue by sharing local observations, rewards, and policies among the agents. Many \emph{centralized training decentralized execution} algorithms are developed based on this idea, such as multi-agent deep deterministic policy gradient (MADDPG) \cite{lowe2020multiagent}, value decomposition network (VDN) \cite{Sunehag2018}, monotonic value function factorization (QMIX) \cite{pmlr-v80-rashid18a}, counterfactual multi agent policy gradient (COMA) \cite{coma}, multi-agent actor-attention-critic (MAAC) \cite{Shariq2019}, multiagent reinforcement learning for unshaped cooperative scenarios (UNMAS) \cite{9525046}, and so on.

Nevertheless, the \emph{centralized training decentralized execution} framework exhibits many limitations. First, the existence of an unavoidable central controller makes the framework vulnerable to single-point failure and cyber-attack. Second, training distributed policies in one single controller can require significant computational and communication resources, raising the issue of scalability and flexibility. Third, collecting agents' local information may raise concerns about personal information leakage, making it difficult to deploy in real-world applications.

The limitations have motivated the design of \emph{fully decentralized} or distributed policy optimization. Distributed policy optimization has attracted significant attention in single-agent RL. For example, \cite{170303864} proposes the use of an evolution strategy (ES) that exploits parallel workers to speed up policy searching, facilitating distributed optimization of neural network policies. \cite{Yang2021NCS} proposes a principled NCS framework that provides theoretical guidance for obtaining optimal NCS processes. However, distributed policy optimization in multi-agent systems faces another challenge - partial observability. Agents may possess incomplete information about the environment state, and to improve observability, many approaches adopt different information structures, such as partially nested \cite{1099850}, periodic sharing \cite{649699}, delayed sharing \cite{5606190}, and event trigger \cite{9597491}. In distributed stochastic control, the common information approach is often employed through partial history sharing \cite{6409396}, and an information state is usually used to represent the agent's posterior belief of the environment state. The sufficiency conditions under which an information state is sufficient are studied in \cite{9360495}. In \cite{9303801}, information state embedding is designed by compressing agents' local histories. However, constructive algorithms are not offered in these studies. In MARL, learnable communication protocols \cite{9781510838819} are preferred, which map local information to actionable messages so that agents can improve observability by incorporating neighbors' messages to augment observations. Various aggregation methods, such as averaging \cite{9781510838819}, encoding \cite{chu2020multiagent}, and attention mechanism \cite{Jiang2018,9716772}, are studied to incorporate neighbors' messages efficiently.

In this paper, we propose a fully decentralized algorithm for partially observable MARL based on TRPO. TRPO has been successful at solving complex single-agent RL problems \cite{pmlr-v37-schulman15} and is known for its effectiveness and monotonic improvement guarantee in optimizing large nonlinear policies \cite{henderson2019deep}. We extend TRPO to cooperative MARL problems where agents have partial observations and private rewards. We introduce the notion of sufficient observation information, and show that the policy update of TRPO can be equivalently transformed into a consensus optimization for networked agents when the local observation is sufficient. We solve the consensus optimization via local linearization, yielding a fully decentralized algorithm, which we call multi-agent TRPO (MATRPO). In MATRPO, distributed policies are optimized using an asynchronous ADMM algorithm, which only requires local communication between neighboring agents. During training, each agent shares the ratio of the local policy to its most recent iterate with neighbors for consensus. Agents do not need to transmit their private information on observations, rewards, or policy and value network parameters. In the experiments, we show that the proposed algorithm can learn high-quality neural network policies for cooperative MARL problems.

\section{Related Work}
There exists a series of insightful works in the literature that address the decentralized MARL problem, and a comprehensive overview of decentralized MARL methods can be found in \cite{zhang2019multiagent,Nguyen2020}. Most decentralized MARL methods fall into two categories: (1) value-base methods \cite{Bertsekas1995}, which alternate between policy evaluation and policy improvement; (2) policy gradient methods \cite{Sutton1999}, which directly optimizes the policy according to an estimated gradient of the expected return.

For value-based methods in decentralized MARL, the key is to evaluate the global value function in a decentralized manner. A distributed policy evaluation algorithm is proposed in \cite{Macua2015} by extending the classic single-agent gradient temporal-difference (GTD) algorithm to multi-agent systems by using diffusion strategies \cite{5475271}. A convergence analysis with linear function approximation under sufficiently small step-size updates was established. Following this line of study, in \cite{7524910} a consensus-based distributed GTD algorithm is studied with a weak convergence analysis under time-varying network topologies using ordinary differential equation \cite{S0363012997331639}. A similar approach formulating the consensus optimization as a primal-dual problem is developed in \cite{Lee2018}. In \cite{Mathkar2017}, an asymptotic convergence of the distributed TD(0) algorithm for both discounted and average reward MARL problems is established considering gossiping agents \cite{1638541}. While the convergence or asymptotic convergence is established in these works, analysis about the rate of convergence or finite-time convergence is missing. In \cite{Hoi2018}, a double averaging scheme combining the dynamic consensus and stochastic average gradient algorithm is proposed with a convergence guarantee at a global geometric rate. In \cite{Doan2019FiniteTimeAO}, the finite-time convergence of a consensus-based distributed TD(0) algorithm under constant and time-varying step-sizes was investigated. Different from the above methods attempting to estimate the value function, the $\mathcal{DQ}$-learning algorithm in \cite{Kar2013} is to learn the action-value function by distributed Q-factors, which were updated by using a $consensus + innovation$  algorithm. However, the aforementioned studies are based on linear function approximation, which limits their application in many complicated MARL problems where general non-linear function approximators, such as neural networks, are required. To address this problem, \cite{Liu2020} presents a neural network based multi-agent deep Q-network (DQN) algorithm, where distributed optimal action-value functions are learned by aggregating neighboring feature sets for state representation via an attentive relational encoder.

Policy gradient methods enable agents to learn stochastic policies, which are essential to partially observable environments \cite{SinghJJ94}. Furthermore, policy gradient methods are more stable in learning complicated nonlinear policies since it can directly search in the space of policy parameters. Most policy gradient methods adopt the actor-critic architecture \cite{Sutton1998}. In \cite{pmlr-v80-zhang18n}, two decentralized actor-critic algorithms are designed by using a consensus action-value function approximator and the policy gradient theorem for decentralized MARL is established. While the proposed method guarantees convergence for linear approximate functions, it does not consider the partial observability issue. In \cite{YZhang2019}, an off-policy actor-critic algorithm is studied by learning consensus policies. By the decomposition of the global value function and action-value function, each agent can estimate the gradient of the consensus-based local policy based on a local critic. The convergence guarantee for linear approximate functions was also established for fully observable environments. However, the proposed method is limited to homogeneous policies since all policies must have the same architecture for consensus. In \cite{9302688}, a semi-centralized MARL algorithm is proposed for cooperative multi-agent StarCraft games. A hierarchical structure with two levels of actor-critic is designed where a global actor-critic learns to provide some global signals based on limited information shared by agents and each agent learns to play using local actor-critics. While this method can reduce the communication burden compared with centralized training methods, it still requires a coordinator simultaneously interacting with all agents during training.

Similar to policy gradient methods, our proposed algorithm can train stochastic policies parameterized by large nonlinear function approximators using local policy search with trust region constraints. Compared to the policy gradient approaches mentioned above, our method considers partial observable environments and is suitable for heterogeneous policies. Specifically, in the proposed algorithm the agents' policies are improved by solving a consensus optimization in a decentralized way using an asynchronous ADMM. Since the consensus is imposed on the policy ratio rather than policy parameters, the policies can have different architectures and the agents only need to share the policy ratio values with neighbors via a peer-to-peer network. Therefore, the proposed algorithm does not require a central controller or a coordinator to collect and distribute every agent's private information on reward, observations, or policy and value function parameters.
\section{Preliminaries}
\label{preliminaries}
\subsection{Partially Observable Markov Game}
We consider a partially observable Markov game (POMG) \cite{Michael1994}, in which $N$ agents communicate with immediate neighbors through a peer-to-peer network to cooperate with each other for a collective goal. The POMG is defined by the tuple $(\mathcal{S},\{\mathcal{O}^{i}\}_{i\in\mathcal{N}},\{\mathcal{A}^{i}\}_{i\in\mathcal{N}}, \mathcal{P},\{\mathcal{P}^{i}_{o}\}_{i\in\mathcal{N}},\{r^{i}\}_{i\in\mathcal{N}},\rho_{0},\gamma,\mathcal{G})$, where $\mathcal{S}$ is a set of environment states, $\mathcal{O}^{i}$ is a set of observations of the agent $i$, $\mathcal{A}^{i}$ is a set of actions that agent $i$ can choose, $\mathcal{P}:\mathcal{S}\times\mathcal{A}^{1}\times\cdots\times\mathcal{A}^{N}\times\mathcal{S}\mapsto\left[0,1\right]$ is the state transition distribution, $\mathcal{P}^{i}_{o}:\mathcal{S}\times\mathcal{O}^{i}\mapsto\left[0,1\right]$ is the distribution of agent $i$'s observation on state $s$, $r^{i}:\mathcal{S}\times\mathcal{A}^{1}\times\cdots\times\mathcal{A}^{N}\mapsto\mathbb{R}$ is the personal reward received by agent $i$, $\rho_{0}:\mathcal{S}\mapsto\left[0,1\right]$ is the distribution of the initial state, $\gamma\in\left[0,1\right)$ is the discount factor, and $\mathcal{G}=(\mathcal{N}, \mathcal{E})$ is an undirected graph representing the communication network, where $\mathcal{N}=\{1,\dots,N\}$ is the set of $N$ agents and $\mathcal{E}=\{1,\dots,L\}$ is the set of $L$ communication links.

Let $\pi^{i}:\mathcal{O}^{i}\times\mathcal{A}^{i}\mapsto\left[0,1\right]$ denote a stochastic policy of agent $i$ and $\pi(a|o)=\prod_{i=1}^{N}\pi^{i}(a^{i}|o^{i})$ the joint policy, where $o=(o^{1},\dots,o^{N})$ is the aggregate observation of all agents and $o^{i}\in\mathcal{O}^{i}$ is the observation of agent $i$; $a=(a^{1},\dots,a^{N})$ is the joint action and $a^{i}\in\mathcal{A}^{i}$ is the action of agent $i$. For the agents, the goal is to learn a set of distributed policy $\{\pi^{i}\}_{i\in\mathcal{N}}$ so that the joint policy $\pi$ maximizes the expected sum of the discounted rewards, $J(\pi)=\mathbb{E}_{\tau\sim\pi}\left[\sum_{t=0}^{\infty}\gamma^{t}(r_{t}^{1}+\cdots+r_{t}^{N})\right]$. Here $\tau$ denotes a trajectory $\tau=(o_{0},a_{1},o_{2},a_{2},\dots)$, and $\tau\sim\pi$ indicates that the distribution over the trajectory depends on $\pi$:
$$s_{0}\sim\rho_{0},\ o_{t}^{i}\sim\mathcal{P}^{i}_{o}(\cdot|s_{t}),\ a_{t}^{i}\sim\pi^{i}(\cdot|o_{t}^{i}),\ s_{t+1}\sim\mathcal{P}(\cdot|s_{t}, a_{t}).$$

Letting $R^{i}(\tau)=\sum_{t=0}^{\infty}\gamma^{t}r_{t}^{i}$ denote the discounted return of a trajectory perceived by agent $i$, we define the local value function as $V^{i}_{\pi}(s)=\mathbb{E}_{\tau\sim\pi}\left[R^{i}(\tau)\mid s_{0}=s\right]$, local action-value function as $Q^{i}_{\pi}(s, a)=\mathbb{E}_{\tau\sim\pi}\left[R^{i}(\tau)\mid s_{0}=s,a_{0}=a\right]$, and local advantage function as $A^{i}_{\pi}(s, a)=Q_{\pi}^{i}(s,a)-V_{\pi}^{i}(s)$, respectively. With these definitions, we can express the global value function by $V_{\pi}(s)=\sum_{i=1}^{N}V_{\pi}^{i}(s)$, the global action-value function by $Q_{\pi}(s,a)=\sum_{i=1}^{N}Q_{\pi}^{i}(s,a)$, and the global advantage function by $A_{\pi}(s,a)=\sum_{i=1}^{N}A_{\pi}^{i}(s,a)$.

Also, we use $\rho_{\pi}$ to denote the discounted state visitation frequencies according to the joint policy $\pi$:
\begin{equation}
\notag
\rho_{\pi}(s)=P(s_{0}=s|\pi)+\gamma P(s_{1}=s|\pi)+\gamma^{2}P(s_{2}=s|\pi)+\cdots.
\end{equation}
Likewise, we use $\rho^{i}_{\pi}$ to denote the discounted observation visitation frequencies of agent $i$ according to $\pi$:
\begin{equation}
\notag
\rho^{i}_{\pi}(o)=P(o^{i}_{0}=o|\pi)+\gamma P(o^{i}_{1}=o|\pi)+\gamma^{2}P(o^{i}_{2}=o|\pi)+\cdots.
\end{equation}
\subsection{Trust Region Policy Optimization}
TRPO is a policy search method with monotonic improvement guarantee \cite{pmlr-v37-schulman15}. In TRPO, the policy is iteratively updated by maximizing a surrogate objective over a trust-region \cite{NoceWrig06,boyd2004convex} around the most recent iterate $\pi_{old}$:
\begin{equation}
\label{standard_trpo}
\begin{aligned}
\max_{\pi}\ \ &\mathbb{E}_{s\sim\rho_{\pi_{old}}, a\sim\pi_{old}}\left[\frac{\pi(a|s)}{\pi_{old}(a|s)}A_{\pi_{old}}(s,a)\right] \\
s.t.\ \ & \overline{D}_{KL}^{\rho_{\pi_{old}}}(\pi_{old},\pi) \leq \delta.
\end{aligned}
\end{equation}
where $\overline{D}_{KL}^{\rho_{\pi_{old}}}(\pi_{old},\pi)=\mathbb{E}_{s\sim\rho_{\pi_{old}}}\left[D_{KL}(\pi_{old}(\cdot|s)\parallel\pi(\cdot|s))\right]$ is the average KL-Divergence. In the TRPO optimization (\ref{standard_trpo}), the objective is an importance sampling \cite{Kahn1951SplittingParticleTransmission} estimator of the advantage function $A_{\pi_{old}}(s,a)$. The KL constraint restricts the searching area in the neighborhood of $\pi_{old}$ via the trust-region parameter $\delta$.

\subsection{Alternating Direction Method of Multipliers}
In our study, we extend the TRPO algorithm to partially observable MARL problems by transforming the optimization in (\ref{standard_trpo}) into a consensus optimization problem (see Section \ref{trpo_decomposition}), which has the following canonical form:
\begin{equation}
\begin{split}
	\min_{x,v}\ &f(x)=\sum_{i=1}^{N}f_{i}(x_{i}) \\
	s.t.\ &x_{i}=z,\ \forall i\in\mathcal{N}
\end{split}
\end{equation}
where the objective function $f$ is \emph{separable} with respect to the decision variables $x=(x_{1},\dots,x_{N})$, i.e., $f(x)=\sum_{i=1}^{N}f_{i}(x_{i})$; $f_{i}$ is the objective function with respect to the local variables $x_{i}\in\mathbb{R}^{m\times 1}$; the constraints $x_{i}=z$ enforce consensus for all $x_{i}$ on $z\in\mathbb{R}^{m\times 1}$. 

ADMM is an effective algorithm to solve problems of such kind with a more general formulation \cite{Stephen2011}:
\begin{equation}
\label{standard_admm}
\begin{aligned}
\min_{x,z}\ &f(x)+g(z) \\
s.t.\ & Ax+Bz=C,
\end{aligned}
\end{equation}
where the matrix $A$ is partitioned to $A=[A_{1},\dots,A_{N}]$ so that $Ax=\sum_{i=1}^{N}A_{i}x^{i}$, $A_{i}\in\mathbb{R}^{p\times m}$, $B\in\mathbb{R}^{p\times m}$, and $C\in\mathbb{R}^{p\times 1}$; the objective is a sum of two convex functions, $f(x)$ and $g(z)$, with respect to the decision variables $x$ and $z$, respectively.

The ADMM algorithm solves the problem (\ref{standard_admm}) by optimizing the following augmented Lagrangian,
\begin{equation}
\label{standard_augmented_lagrange}
\begin{aligned}
\mathcal{L}_{\beta}(x,z,y)=&f(x)+g(z)+y^{T}(Ax+Bz-C)\\
&+\frac{\beta}{2}\parallel Ax+Bz-C\parallel_{2}^{2},\ \beta>0,
\end{aligned}
\end{equation}
where $y\in\mathbb{R}^{m\times 1}$ is the dual variable or Lagrange multiplier,. The optimization is performed through sequentially updating the primal variables $x$ and $z$, and the dual variable $y$:
\begin{subequations}
\label{standard_ADMM}
\begin{align}
&x^{(k+1)}_{i}:=\argmin_{x_{i}}\mathcal{L}_{\rho}(x_{i},z^{(k)},y^{(k)}),\ \forall i \\
&z^{(k+1)}:=\argmin_{z}\mathcal{L}_{\rho}(x^{(k+1)},z,y^{(k)}), \\
&y^{(k+1)}:=y^{(k)}+\beta(Ax^{(k+1)}+Bz^{(k+1)}-C).
\end{align}
\end{subequations}

\section{Multi-Agent Trust Region Policy Optimization}
\subsection{Trust Region Optimization for Multiple Agents}
\label{trpo_decomposition}
Consider the TRPO policy optimization in (\ref{standard_trpo}) for the multi-agent case. We decompose the advantage function $A_{\pi_{old}}(s,a)$ in the objective and rewrite it as follows:
\begin{equation}
\label{decomposed_trpo}
\begin{split}
\max_{\pi}&\underset{\substack{s\sim\rho_{\pi_{old}}\\ a\sim\pi_{old}}}{\mathbb{E}}\left[\frac{\pi(a|s)}{\pi_{old}(a|s)}\left(A^{1}_{\pi_{old}}(s,a)+\cdots+A^{N}_{\pi_{old}}(s,a)\right)\right] \\
s.t.&\ \overline{D}_{KL}^{\rho_{\pi_{old}}}(\pi_{old},\pi) \leq \delta.
\end{split}
\raisetag{1.2\normalbaselineskip}
\end{equation}
The purpose is to split the objective into $N$ independent sub-objectives so that the optimization problem can be solved in a distributed way by $N$ agents. However, the sub-objectives are coupled through the joint policy ratio $\pi(a|s)/\pi_{old}(a|s)$, where $\pi(a|s)=\prod_{i=1}^{N}\pi^{i}(a^{i}|s)$.

To solve this problem, we let every agent learn a local model of the joint policy, $\pi^{i}(a|s), i\in\mathcal{N}$ and reach an agreement on the policy ratio $\pi^{i}(a|s)/\pi^{i}_{old}(a|s)$. Then, the policy optimization (\ref{decomposed_trpo}) can be equivalently transformed into the the consensus optimization problem (see Appendix \ref{appendix-A}):
\begin{equation}
\label{matrpo}
\begin{split}
\max_{\{\pi^{i}\}_{i\in\mathcal{N}}}\ \ &\sum_{i=1}^{N}\underset{\substack{s\sim\rho_{\pi_{old}}\\ a\sim\pi_{old}}}{\mathbb{E}}\left[\frac{\pi^{i}(a|s)}{\pi^{i}_{old}(a|s)}A^{i}_{\pi_{old}}(s,a)\right] \\
s.t.\ \ & \frac{\pi^{1}(a^{i}|s)}{\pi^{1}_{old}(a^{i}|s)}=\cdots=\frac{\pi^{N}(a^{i}|s)}{\pi^{N}_{old}(a^{i}|s)}, \forall i\in\mathcal{N}\\
& \overline{D}_{KL}^{\rho_{\pi_{old}}}(\pi_{old},\pi) \leq \delta.
\end{split}
\raisetag{3.2\normalbaselineskip}
\end{equation}

Now, the objective is separable with respect to $\{\pi^{i}(a|s)\}_{i\in\mathcal{N}}$. However, the local model $\pi^{i}(a|s)$ is conditioned on the global $s$, which is unavailable in a POMG. Moreover, the KL constraint is imposed on the joint policy $\pi_{old}$, which is difficult to satisfy in a distributed way. To address this problem, we introduce the following definition and theorem.
\begin{definition}[Sufficient Observation Information]
	The local observation $o^{i}$ of agent $i$ is said to be sufficient if it satisfies:
\begin{equation}
\notag
\begin{split}
	&\textbf{\textup{(1)}}\ \mathbb{E}_{o^{i}\sim\mathcal{P}^{i}_{o}(\cdot|s),a\sim\pi}[r^{i}(s,a)|s]=\mathbb{E}_{s\sim\mathcal{P}_{s}(\cdot|o^{i}),a\sim\pi}[r^{i}(s,a)|o^{i}],\\
	&\textbf{\textup{(2)}}\ \mathcal{P}_{\pi}(s',o^{i'}|s)=\mathcal{P}_{\pi}(s',o^{i'}|o^{i}),
\end{split}
\end{equation}
where $\mathcal{P}_{\pi}(s',o^{i'}|s)$ and $\mathcal{P}_{\pi}(s',o^{i'}|o^{i})$ are the transition probabilities from $s$ and $o^{i}$ to $(s',o^{i'})$ under $\pi$, respectively.
\end{definition}
\begin{theorem}
\label{main_result}
If the local observations $o^{i},\forall i\in\mathcal{N}$ is sufficient, then the problem (\ref{pomatrpo}) is equivalent to the problem (\ref{decomposed_trpo}) with a stricter trust region, i.e., $\overline{D}_{KL}^{\rho_{\pi_{old}}}(\pi_{old},\pi) \leq \Delta$, where $\Delta\leq\delta$.
\begin{equation}
\label{pomatrpo}
\begin{aligned}
\max_{\{\pi^{i}\}_{i\in\mathcal{N}}}\ \ &\sum_{i=1}^{N}\underset{\substack{o^{i}\sim\rho_{\pi_{old}}^{i}\\ a\sim\pi_{old}}}{\mathbb{E}}\left[\frac{\pi^{i}(a|o^{i})}{\pi^{i}_{old}(a|o^{i})}A^{i}_{\pi_{old}}(o^{i},a)\right] \\
s.t.\ \ & \frac{\pi^{1}(a^{i}|o^{1})}{\pi^{1}_{old}(a^{i}|o^{1})}=\cdots=\frac{\pi^{N}(a^{i}|o^{N})}{\pi^{N}_{old}(a^{i}|o^{N})}, \forall i\in\mathcal{N} \\
& \overline{D}_{KL}^{\rho^{i}_{\pi_{old}}}(\pi^{i}_{old},\pi^{i}) \leq \frac{\delta}{N}, \forall i\in\mathcal{N},
\end{aligned}
\end{equation}
where $\pi^{i}(a|o^{i})=\prod_{n=1}^{N}\pi^{i}(a^{n}|o^{i})$, and $a\sim\pi_{old}$ means that $a$ is selected according to the joint polity $\pi_{old}$; the KL Divergence is $\overline{D}_{KL}^{\rho^{i}_{\pi_{old}}}(\pi^{i}_{old},\pi^{i})=\mathbb{E}_{o^{i}\sim\rho^{i}_{\pi_{old}}}\left[D_{KL}(\pi^{i}_{old}(\cdot|o^{i})\parallel\pi^{i}(\cdot|o^{i}))\right].$
\end{theorem}
\begin{proof}[Proof]
The proof is provided in Appendix \ref{appendix-A}.
\end{proof}
Theorem \ref{main_result} provides a theoretical foundation for extending TRPO to POMG problems. It means that the policy optimization in TRPO can be transformed equivalently into a consensus optimization if the agents have sufficient observation. Various approaches can be used to improve observability, such as extracting information from histories. This is beyond the scope of this paper and we refer interested readers to \cite{6409396,chu2020multiagent}.
\subsection{Distributed Consensus with Asynchronous ADMM}
There are many techniques that can solve distributed consensus problems, such as gossip algorithms \cite{1638541}, distributed gradient methods \cite{4749425}, and ADMM \cite{Stephen2011}. Gossip algorithms generally focus on the consensus averaging problem for a network of computational nodes. Distributed gradient methods combine consensus averaging with local gradient descent steps and usually apply to unconstrained problems. ADMM adopts primal-dual formulation for constrained distributed optimization problems, and it is suitable for our problem.

Traditional distributed ADMM algorithms generally require a synchronous process at each iteration. As shown in Eq. (\ref{standard_ADMM}), to optimize the primal variables $x_i$, the values of $z^{(k)}$ and $y^{(k)}$ at the most recent update need to be broadcasted. Then, the new primal variables $x^{(k+1)}=(x_{1}^{(k+1)},\dots,x_{N}^{(k+1)})$ must be synchronized to update the variables $z$ and $y$. This can increase the communication burden in a large-scale network and slow down the pace of the optimization.

In our study, we employ the asynchronous distributed ADMM algorithm \cite{Wei2013}. The algorithm allows asynchronous updates by exchanging information between randomly activated neighboring agents. Moreover, the algorithm is robust to random failure of the communications. These advantages 
Next, we describe the asynchronous distributed ADMM algorithm.

\begin{figure}
\centering     
\includegraphics[width=70mm]{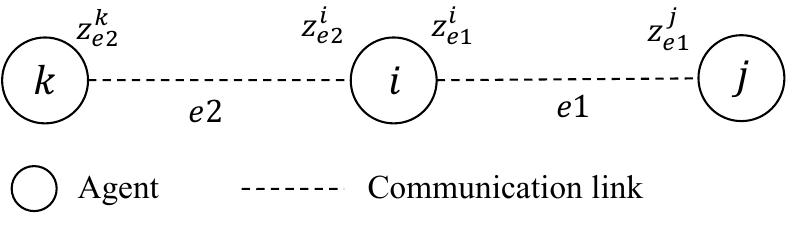}
\caption[Caption for LOF]{For the communication link $e1$, the endpoints are agents $i$ and $j$, and we have $\mathcal{N}(e1)=\{i,j\}$. In the asynchronous distributed ADMM \cite{Wei2013}, $z_{e1}^{i}$ is used by agent $i$ to estimate the policy ratio of agent $j$. Since agent $i$ also connects to agent $k$ via the link $e2$, agent $i$ uses another variables $z_{e2}^{i}$ to estimate the policy ratio of agent $k$.}
\label{admm}
\end{figure}

Let $\mathcal{N}(e)=\{i,j\}$ denote the agents at the endpoints of the communication link $e\in\mathcal{E}$. For each agent $q\in\{i,j\}$, we introduce the variable $z_{e}^{q}(o^{q},a^{n}), \forall n\in\mathcal{N}$ to estimate the likelihood ratio of its neighbor $\mathcal{N}(e)\backslash\{q\}$. An example of $z_{e}^{q}$ is illurstrated in Fig. \ref{admm}. The consensus constraint for the agent $i$ and $j$ can then be replaced by:
\begin{subequations}
\label{admm_constraint}
\begin{align}
&\frac{\pi^{i}(a^{n}|o^{i})}{\pi^{i}_{old}(a^{n}|o^{i})}=z_{e}^{i}(o^{i},a^{n}),\ -\frac{\pi^{j}(a^{n}|o^{j})}{\pi^{j}_{old}(a^{n}|o^{j})}=z_{e}^{j}(o^{j},a^{n}), \\
&z_{e}^{i}(o^{i},a^{n})+z_{e}^{j}(o^{j},a^{n})=0.
\end{align}
\end{subequations}
Let the constant $C_{e}^{q}$ be the weight on the information transmitted between agents $q\in\mathcal{N}(e)$. $C_{e}^{q}$ takes the value of $1$ or $-1$ such that $C_{e}^{i}+C_{e}^{j}=0$.
Combining Eqs. (\ref{admm_constraint}), the problem (\ref{pomatrpo}) can be transformed into
\begin{equation}
\label{asyadmm}
\begin{split}
\underset{\substack{\pi^{i},z^{q}_{e}}}{\min}\ &-\sum_{i=1}^{N}\underset{\substack{o^{i}\sim\rho_{\pi_{old}}^{i}\\ a\sim\pi_{old}}}{\mathbb{E}}\left[\frac{\pi^{i}(a|o^{i})}{\pi^{i}_{old}(a|o^{i})}A^{i}_{\pi_{old}}(o^{i},a)\right] \\
s.t.\ &\ C_{e}^{q}\ \frac{\pi^{i}(a^{n}|o^{i})}{\pi^{i}_{old}(a^{n}|o^{i})}=z_{e}^{q}(o^{q},a^{n}),\ q\in\mathcal{N}(e), \\
\ &z_{e}^{i}(o^{i},a^{n})+z_{e}^{j}(o^{j},a^{n})=0,\ (i,j)\in\mathcal{N}(e), \\
\ &\overline{D}_{KL}^{\rho_{\pi_{old}}^{i}}(\pi_{old}^{i},\pi^{i}) < \delta/N,\ \forall i\in\mathcal{N}.
\end{split}
\end{equation}

Note that $\pi^{i}(a^{n}|o^{i})$ and $z_{e}^{q}(o^{q},a^{n})$ are dependent on $o^{i}$ and $a^{n}$. For notational simplicity, we denote $\pi^{i}$ as a vector of the decision variables $\pi^{i}(a|o^{i})$ for all realizations of $o^{i}$ and $a^{n}$. Similarly, we denote $z_{e}^{qn}$ as a vector of the decision variables $z_{e}^{q}(o^{q},a^{n})$ for all realizations of $o^{i}$ and $a^{n}$. We also use the notations $\Upsilon^{n}(\pi^{i}):=\pi^{i}(a^{n}|o^{i})/\pi^{i}_{old}(a^{n}|o^{i})$ and $\Upsilon(\pi^{i}):=\pi^{i}(a|o^{i})/\pi^{i}_{old}(a|o^{i})$ to represent the policy ratios for all realizations of $o^{i}$ and $a^{n}$. Then, we form the augmented Lagrangian of (\ref{asyadmm}):
\begin{equation}
\label{asy_admm_augmented_lagrange}
\begin{split}
\mathcal{L}_{\beta}&(\bm{\pi},\bm{z},\bm{y})=-\sum_{i=1}^{N}\mathbb{E}_{o^{i}\sim\rho_{\pi_{old}}^{i},a\sim\pi_{\tiny old}}\left[\Upsilon(\pi^{i})A^{i}_{\pi_{old}}(o^{i},a)\right] \\
&+\sum_{e\in\mathcal{E}}\sum_{q\in\mathcal{N}(e)}\sum_{n\in\mathcal{N}}y_{e}^{qn}\left(C_{e}^{q}\Upsilon^{n}(\pi^{q})-z_{e}^{qn}\right)\\
&+\sum_{e\in\mathcal{E}}\sum_{q\in\mathcal{N}(e)}\sum_{n\in\mathcal{N}}\frac{\beta}{2}\parallel C_{e}^{q}\Upsilon^{n}(\pi^{q})-z_{e}^{qn}\parallel_{2}^{2},
\end{split}
\raisetag{1.7\normalbaselineskip}
\end{equation}
where $\bm{\pi}=\{\pi^{i}\}_{i\in\mathcal{N}}$ and $\bm{z}=\{z_{e}^{qn}\}_{e\in\mathcal{E},q\in\mathcal{N}(e),n\in\mathcal{N}}$ are primal variables and $\bm{y}=\{y_{e}^{qn}\}_{e\in\mathcal{E},q\in\mathcal{N}(e),n\in\mathcal{N}}$ are dual variables; $\beta>0$ is a penalty parameter. The primal variables $\pi^{i}$ and $z_{e}^{qn}$ satisfy the constraints:
$\Pi^{i}=\{\pi\vert\overline{D}_{KL}^{\rho_{\pi_{old}}^{i}}(\pi_{old}^{i},\pi) < \delta/N\}$ and
$Z_{e}^{n}=\{z_{e}^{in},z_{e}^{jn}\vert z_{e}^{in}+z_{e}^{jn}=0, i,j\in\mathcal{N}(e)\}.$

To solve problem (\ref{asyadmm}), the asynchronous ADMM minimizes the augmented Lagrangian by sequentially updating the local policies $\bm{\pi}$, the estimators $\bm{z}$, and the dual variables $\bm{y}$. Specifically, let $\bm{\pi}^{(k)}$, $\bm{z}^{(k)}$, $\bm{y}^{(k)}$ be the values at iteration $k$. At iteration $k+1$, a communication link $e=(i,j)$ is selected and the two agents $q\in\mathcal{N}(e)$ at the endpoints are activated to update $\pi^{q}$, $z_{e}^{qn}$, and $y_{e}^{qn}$ according to:
\begin{subequations}
\label{asyn_admm}
\begin{align}
&\pi^{q(k+1)} :=\argmin_{\pi^{q}\in\Pi^{q}}\mathcal{L}_{\beta}(\bm{\pi},\bm{z}^{(k)},\bm{y}^{(k)}), \\
&z^{qn(k+1)}_{e} :=\argmin_{z^{qn}_{e}\in Z_{e}^{n}}\mathcal{L}_{\beta}(\bm{\pi}^{(k+1)},\bm{z},\bm{y}^{(k)}), \\
&y^{qn(k+1)}_{e}:=y^{qn(k)}_{e}+\beta(C_{e}^{q}\Upsilon^{n}(\pi^{q(k+1)})-z^{qn(k+1)}_{e}).
\raisetag{3.5\normalbaselineskip}
\end{align}
\end{subequations}

The algorithm (\ref{asyn_admm}) converges almost surely to the optimal solution at the rate of $\mathcal{O}(1/k)$ as $k\mapsto\infty$ if the objectives and the sets $\Pi^{i}$ and $Z_{e}^{n}$ are convex \cite{Wei2013}. However, the convex assumption does not hold for $\Pi^{i}$ due to the KL constraints. Besides, we may need to use parameterized policies $\pi^{i}(a|o^{i};\theta^i)$ when the observation and action space is huge. For general nonlinear policies, like neural networks, the convex assumption does not hold, and the convergence of the asynchronous ADMM algorithm in (\ref{asyn_admm}) is not guaranteed \cite{Stephen2011, Wei2013}.
\subsection{Sequential Convexification for Convergence}
To address the convergence issue, we borrow the idea from sequential convex programming \cite{Magnusson2015} and approximate the problem (\ref{asyadmm}) with a convex model. The idea is also adopted in TRPO \cite{pmlr-v37-schulman15} and constrained policy optimization (CPO) \cite{Joshua2017}. Since we are considering parameterized policy $\pi^{i}(a|o^{i};\theta^{i})$, we will use the parameter vector $\theta^{i}$ to represent the policy $\pi^{i}$, and overload all previous notations, e.g. $\Upsilon^{n}(\theta^{i}):=\Upsilon^{n}(\pi^{i})$, $f^{i}(\pi^{i}):=f^{i}(\theta^{i})$. Note that, for a small trust region $\delta$, the likelihood ratio $\Upsilon^{n}(\theta^{i})$ and $\Upsilon(\theta^{i})$ can be well approximated by first order Taylor expansion round $\theta^{i}_{old}$:
\begin{equation}
\begin{split}
\label{appr_ratio}
\Upsilon^{n}(\theta^{i}) &\approx 1 + \nabla_{\theta^{i}}^{T}\Upsilon^{n}(\theta^{i})(\theta^{i}-\theta^{i}_{old}), \\
\Upsilon(\theta^{i}) &\approx 1 + \textstyle\sum_{n=1}^{N}\nabla_{\theta^{i}}^{T}\Upsilon^{n}(\theta^{i})(\theta^{i}-\theta^{i}_{old})
\end{split}
\end{equation}
and the KL-divergence can be well approximate by the second-order Taylor expansion round $\theta^{i}_{old}$:
\begin{equation}
\label{appr_kl}
\overline{D}_{KL}^{\rho_{\theta_{old}}^{i}}(\theta^{i}_{old},\theta^{i}) \approx \frac{1}{2}(\theta^{i}-\theta^{i}_{old})^{T}H_{i}(\theta^{i}-\theta^{i}_{old}),
\end{equation}
where $H_{i}=\nabla_{\theta^{i}}^{2}\overline{D}_{KL}^{\rho_{\theta_{old}}^{i}}(\theta^{i}_{old},\theta^{i})$. Substituting (\ref{appr_ratio}) - (\ref{appr_kl}) into (\ref{asyadmm}) and (\ref{asy_admm_augmented_lagrange}), we can derive a convex approximation of the primal problem:
\begin{equation}
\label{convex_asyadmm}
\begin{split}
\underset{\substack{\theta^{i},z_{e}^{q}}}{\min}\ &-\sum_{i=1}^{N}\sum_{n=1}^{N}\overline{A}^{i}_{n}(\theta_{old}^{i})(\theta^{i}-\theta^{i}_{old}) \\
s.t.\ &\ C_{e}^{q}+C_{e}^{q}\nabla_{\theta^{i}}^{T}\Upsilon^{n}(\theta^{i})(\theta^{i}-\theta^{i}_{old})=z_{e}^{q},\ q\in\mathcal{N}(e), \\
\ &z_{e}^{in}+z_{e}^{jn}=0,\ (i,j)\in\mathcal{N}(e), \\
\ &\frac{1}{2}(\theta^{i}-\theta^{i}_{old})^{T}H_{i}(\theta^{i}-\theta^{i}_{old}) < \delta/N,\ \forall i\in\mathcal{N}, \\
\ & e\in{\mathcal{E}}, n\in\mathcal{N}.
\end{split}
\raisetag{1.3\normalbaselineskip}
\end{equation}
where $\overline{A}^{i}_{n}(\theta_{old}^{i})=\mathbb{E}_{o^{i}\sim\rho_{\theta_{old}}^{i},a\sim\theta_{old}}\left[A^{i}_{\theta_{old}}(o^{i},a)\nabla_{\theta^{i}}\Upsilon^{n}(\theta^{i})\right]$. The corresponding augmented Lagrangian is
\begin{equation}
\label{convex_asy_admm_augmented_lagrange}
\begin{split}
&\mathcal{L}_{\beta}(\bm{\theta},\bm{z},\bm{y})=-\sum_{i=1}^{N}\sum_{n=1}^{N}\overline{A}^{i}_{n}(\theta_{old}^{i})(\theta^{i}-\theta^{i}_{old}) \\
&+\sum_{e\in\mathcal{E}}\sum_{q\in\mathcal{N}(e)}\sum_{n\in\mathcal{N}}y_{e}^{qn}\left(C_{e}^{q}+C_{e}^{q}\nabla_{\theta^{i}}^{T}\Upsilon^{n}(\theta^{i})(\theta^{i}-\theta^{i}_{old})-z_{e}^{qn}\right)\\
&+\sum_{e\in\mathcal{E}}\sum_{q\in\mathcal{N}(e)}\sum_{n\in\mathcal{N}}\frac{\beta}{2}||C_{e}^{q}+C_{e}^{q}\nabla_{\theta^{i}}^{T}\Upsilon^{n}(\theta^{i})(\theta^{i}-\theta^{i}_{old})-z_{e}^{qn}||_{2}^{2},
\end{split}
\raisetag{6.8\normalbaselineskip}
\end{equation}
where the decision variables $\theta^{i}$ are restricted to:
$\Theta^{i}=\{\frac{1}{2}(\theta^{i}-\theta^{i}_{old})^{T}H_{i}(\theta^{i}-\theta^{i}_{old}) < \delta/N\}, i\in\mathcal{N}.$ Then, the asynchronous ADMM algorithm becomes
\begin{subequations}
\label{convex_asyn_admm}
\begin{align}
\theta^{q(k+1)} :=&\argmin_{\theta^{q}\in\Theta^{q}}\mathcal{L}_{\beta}(\bm{\theta},\bm{z}^{(k)},\bm{y}^{(k)}), \\
z^{qn(k+1)}_{e} :=&\argmin_{z^{qn}_{e}\in Z_{e}^{n}}\mathcal{L}_{\beta}(\bm{\theta}^{(k+1)},\bm{z},\bm{y}^{(k)}), \\
y^{qn(k+1)}_{e} :=&\ y^{qn(k)}_{e}+\beta\big[C_{e}^{q}+C_{e}^{q}\nabla_{\theta^{i(k+1)}}^{T}\Upsilon^{n}(\theta^{i(k+1)}) \\
\notag&(\theta^{i(k+1)}-\theta^{i(k+1)}_{old})-z_{e}^{qn(k+1)}\big].
\end{align}
\end{subequations}
Due to the convex approximation, the algorithm may converge to stationarity instead of optimality. However, since deep neural networks are highly nonlinear and non-convex, it is difficult for the algorithm to surely converge to optimality. Despite this limitation, the proposed approach demonstrates good performance in practice as shown in the experiments.

\subsection{Information Required by Agents}
To optimize the local policy, each agent needs to compute the local advantage $A^{i}_{\theta_{old}}(o^{i},a)$ and the policy ratio $\Upsilon^{n}(\theta^{i}), \forall n\in\mathcal{N}$. To that end, each agent will need the local observation $o^{i}$, the joint action $a=(a^{1},\dots,a^{N})$, and the personal reward $r^{i}$. The joint action $a$ is only global information required by each agent. Being able to observe other agents' actions is not a restrictive assumption in many cooperative tasks \cite{pmlr-v80-zhang18n,YZhang2019}. However, it is worth mentioning that knowing the actions of other agents does not mean knowing their policies because the local policy $\pi^{i}_{old}(a|o^{i})$ is conditioned on local observation $o^{i}$, which is owned solely by agent $i$.
\section{Practical Implementation}
In this section, we present the practical implement of the MATRPO algorithms for problems where the observation space and action space are large. First, we estimate the expected advantage function and the consensus constraint based on samples. Second, we show that we can replace the policy ratio in the consensus constraint with the logarithmic policy ratio, which enables the development of a closed-form solution.
\subsection{Sample-Based Approximation of the Consensus Constraint}
\label{sample_approx}
To satisfy the consensus constraint, the agents need to reach a consensus on the policy ratio $\Upsilon^{n}(\theta^{i})=\frac{\pi^{i}(a^{n}|o^{i};\theta^{i})}{\pi^{i}(a^{n}|o^{i};\theta^{i}_{old})}, \forall n\in \mathcal{N}$ for every observation and action. This is challenging when the observation and action spaces are large. In practice, the consensus constraints are implemented based on samples. Specifically, we simulate the joint policy $\pi(a|o)$ for $T$ steps in an environment to generate a batch of trajectories of the states and actions, and each agent $i$ has a batch of $D$ locally observed trajectories, i.e. $\mathcal{D}^{i}=\{(o^{i}_{0,d},a_{0,d},o^{i}_{1,d},\dots,a_{T-1,d},o^{i}_{T-1,d})_{d=1,\dots,D}\}$. After that, the agents compute the policy ratios $\Upsilon^{n}(\theta^{i})$ and communicate with their neighbors. Then, we constrain the agents to reach consensus on these sampled policy ratios.

For the calculation of local advantage function $A^{i}_{\pi_{old}}(o^{i},a)$, we use the generalized advantage estimation (GAE) method \cite{SchulmanMLJA15}:
\begin{equation}
A^{i}=\epsilon^{i}_{t}+(\gamma\lambda)\epsilon^{i}_{t+1}+\cdots+(\gamma\lambda)^{T-t+1}\epsilon^{i}_{T-1},
\end{equation}
\begin{equation}
	\epsilon^{i}_{t}=r^{i}_{t}+\gamma V^{i}_{\pi_{old}}(o^{i}_{t+1}) - V^{i}_{\pi_{old}}(o^{i}_{t}),
\end{equation}
where $V^{i}_{\pi_{old}}(o^{i})$ is the local value function approximated by a neural network.
\subsection{Agreement on Logarithmic Ratio}
We find that it is better to use the logarithmic likelihood ratio $\log(\Upsilon^{n}(\theta^{i}))$ for consensus since it leads to a closed-form solution to the asynchronous ADMM updates (\ref{asyn_admm}). As the logarithmic function is monotonic, consensus on $\log(\Upsilon^{n}(\theta^{i}))$ is equivalent to consensus on $\Upsilon^{n}(\theta^{i})$. 
In addition, the first order Taylor expansion of the logarithmic policy ratio is the same as that of the policy ratio:  
\begin{equation}
\label{appr_log_ratio}
\nabla_{\theta^{i}}^{T}\log(\Upsilon^{n}(\theta^{i}))(\theta^{i}-\theta^{i}_{old}) =\nabla_{\theta^{i}}^{T}\Upsilon^{n}(\theta^{i})(\theta^{i}-\theta^{i}_{old}).
\end{equation}
Then, we can approximate Eq. (\ref{convex_asy_admm_augmented_lagrange}) by the following Lagrangian based on the Monte-Carlo sampling method \cite{DKTTZBWiley11}:
\begin{equation}
\label{sample_augmented_lagrange}
\begin{split}
\mathcal{L}_{\beta}&(\bm{\theta},\bm{z},\bm{y})=-\frac{1}{M}\sum_{i=1}^{N}\sum_{n=1}^{N}{A^{i}}^{T} J^{in}(\theta^{i}-\theta^{i}_{old}) \\
&+\frac{1}{M}\sum_{e=1}^{L}\sum_{q\in\mathcal{N}(e)}\sum_{n=1}^{N}{y_{e}^{qn}}^{T}\left[C_{e}^{q}J^{qn}(\theta^{q}-\theta^{q}_{old})-z_{e}^{qn}\right] \\
&+\frac{1}{M}\sum_{e=1}^{L}\sum_{q\in\mathcal{N}(e)}\sum_{n=1}^{N}\frac{\beta}{2}\parallel C_{e}^{q}J^{qn}(\theta^{q}-\theta^{q}_{old})-z_{e}^{qn}\parallel_{2}^{2},
\end{split}
\raisetag{8.1\normalbaselineskip}
\end{equation}
where $A^{i}\in\mathbb{R}^{M\times 1}$ is a vector of the sampling values of the advantage function $A^{i}_{\pi_{old}}(o^{i},a)$, $J^{in}\in\mathbb{R}^{M\times D}$ is the Jacobian of the sampled likelihood ratios $\Upsilon^{n}(\theta^{i})$ with respect to $\theta^{i}\in\mathbb{R}^{D\times 1}$. The policy parameters $\theta^{i}$ is defined in the feasible set $\theta^{i}\in\Theta^{i}=\{\theta\vert\frac{1}{2}(\theta-\theta^{i}_{old})^{T}\overline{H}^{i}(\theta-\theta^{i}_{old})\leq\delta/N\}$, where $\overline{H}^{i}\in\mathbb{R}^{D\times D}$ is the Hessian of the sample average KL-Divergence $\frac{1}{M}\sum_{m=1}^{M}D_{KL}(\pi^{i}_{old}(\cdot|o^{i};\theta^{i}_{old})||\pi^{i}(\cdot|o^{i};\theta^{i}))$.

The asynchronous ADMM updates for the Lagrangian (\ref{sample_augmented_lagrange}) have the following closed-forms (see Appendix \ref{appendix-B} for proof):
\begin{equation}
\label{log_appr_asy_admm_updates}
\begin{split}
\theta^{q(k+1)}&:=\theta^{q}_{old}+\sqrt{\frac{2\delta/N}{V^{T}{\overline{H}^{q}}^{\ -1}V}}{\overline{H}^{q}}^{\ -1}V, \\
z_{e}^{qn(k+1)}&:=\ \frac{1}{\beta}(y_{e}^{qn(k)}-\nu_{e}^{n})+C_{e}^{q}J^{qn}(\theta^{q(k+1)}-\theta^{q}_{old}), \\
y_{e}^{qn(k+1)}&:=\ \nu_{e}^{n},
\end{split}
\raisetag{1.2\normalbaselineskip}
\end{equation}
where
\begin{equation}
\notag
V=\frac{1}{M}\sum_{n=1}^{N}{J^{qn}}^{T}(A^{q}-\sum_{e\in\mathcal{E}(q)}C_{e}^{q}y_{e}^{qn(k)}+\beta\sum_{e\in\mathcal{E}(q)} C_{e}^{q}z_{e}^{qn(k)}),
\end{equation}
\begin{equation}
\begin{split}
\notag
\nu_{e}^{n}=&\frac{1}{2}\sum_{q\in\mathcal{N}(e)}\left[y_{e}^{qn(k)}+\beta C_{e}^{q}J^{qn}(\theta^{q(k+1)}-\theta^{q}_{old})\right].
\end{split}
\end{equation}
and $\mathcal{E}(q)$ is the set of all communication links that directly connect to agent $q$. The pseudocode for our algorithm is given as Algorithm \ref{alg:matrpo}.
\begin{algorithm}[tb]
   \caption{Multi-Agent Trust Region Policy Optimization
   }
   \label{alg:matrpo}
\begin{algorithmic}
   \STATE {\bfseries Input:} Initial local policies $\pi^{i}(\theta^{i}_{0}),\forall i\in\mathcal{N}$; tolerance $\delta$
   \FOR{$itr=0,1,2,\dots$}
   \STATE Set the joint policy $\pi(\theta_{old})=\prod_{i=1}^{N}\pi^{i}(a^{i}|o^{i};\theta^{i}_{itr})$
   \FOR{$i=0,1,\dots,N$}
   \STATE Sample a set of trajectories $\mathcal{D}^{i}\sim\pi(\theta_{old})$ for agent $i$
   \STATE Compute $A^{i}, J^{in},\overline{H}^{i}, \forall i,n\in\mathcal{N}$ using samples 
   \ENDFOR
   \STATE Initial estimators $\bm{z}^{(0)}$ and dual variables $\bm{y}^{(0)}$
   \FOR{$k=0,1,2,\dots$}
   \STATE Randomly select a communication link $e$
   \STATE For agents $q:=(i,j)\in\mathcal{N}(e)$ at the endpoints of link $e$, update $\theta^{q}$, $z^{qn}_{e}$ and $y^{qn}_{e}$ according to (\ref{log_appr_asy_admm_updates})
   \STATE Update $\bm{\theta}^{(k)},\bm{z}^{(k)},\bm{y}^{(k)}\rightarrow \bm{\theta}^{(k+1)},\bm{z}^{(k+1)},\bm{y}^{(k+1)}$
   \ENDFOR
   \ENDFOR
\end{algorithmic}
\end{algorithm}
\section{Empirical Study}
In this section, we design experiments to test the proposed MATRPO algorithm. Through experimental studies, we would like to answer the following questions:
\begin{itemize}
  \item Does MATRPO succeed in learning cooperative policies in a fully decentralized manner when agents have different observation spaces and reward functions?
  \item Is MATRPO comparable with other MARL approaches, such as centralized training and decentralized execution algorithms? Does it outperform independent learning? 
  \item Can MATRPO be used to solve large-scale MARL problems? Does the performance degrade when the number of agents increases?
\end{itemize}

We first conduct experiments on the Multi-agent Particle Environment \cite{mordatch2018emergence} on two tasks: 1) Cooperative Navigation, and 2) Cooperative Treasure Collection. Both tasks require multiple agents to collaborate with partial observations and individual rewards. Finally, we evaluate MATRPO and the baselines on the StarCraft Multi-Agent Challenge \cite{samvelyan19smac}, which focuses on decentralized micromanagement of a group of agents trying to defeat enemies controlled by carefully handcrafted heuristics.

\subsection{Baselines}
\emph{I-TRPO}: An independent learning algorithm where the agents learn individual policies based on TRPO in the single-agent settings. Specifically, a policy $\pi^{i}(a^{i}|o^{i})$ is learned by a neural network for each agent $i$, which maximizes the expected return based on the local observation $o^{i}$. Additionally, a value network is trained for policy evaluation using $o^{i}$ and the private reward $r^{i}$.

\emph{FDMARL}: A fully decentralized algorithm \cite{pmlr-v80-zhang18n} based on the actor-critic design. In the algorithm, every agent learns a policy network and a consensus critic network. The policy network is trained by using typical policy gradient. The critic network is trained based on the action-value TD-learning followed by a diffusion update, where a linear combination of its neighbors' parameter estimates are used for the agents to reach a consensus on the critic network parameters.  

\emph{MAPPO}: A centralized training and decentralized execution algorithm proposed in \cite{yu2022the}, which resembles the structure of PPO by learning a policy network and a value network for each agent. The value network is trained by using the global information on shared observations and shared rewards. The policy network is trained by using the agent's local observations based on PPO.

In all experiments, we use feedforward neural networks with two hidden layers of 128 SeLU units \cite{NIPS2017Gunter} for the policy and value networks. The decentralized communication network has a ring topology. For the implementations of FDMARL \cite{pmlr-v80-zhang18n} and MAPPO \cite{yu2022the}, we use the recommended settings of the hyper-parameters. We run all experiments 5 times with different random seeds. The simulations are carried out on a computer with an Intel Core i7-7700X Processor 3.60GHz and a 16 GB memory. The operation system is Ubuntu 20.04.3 LTS. The code is written in Python 3.7.6 using the deep learning toolbox TensorFlow 2.2.0 \cite{tensorflow2015-whitepaper}, and the RL toolboxes OpenAI Gym \cite{gym} and Baselines-tf2 \cite{baselines}.
\subsection{Multi-Agent Particle Environment (MPE)}
\subsubsection{Environment Description}
MPE is first developed in \cite{mordatch2018emergence} for MARL research, and extended by \cite{lowe2020multiagent} and \cite{Shariq2019}. We test the proposed algorithm on two cooperative tasks (Fig.~\ref{icml-envs}):

\begin{figure}
\centering     
\subfigure[Cooperative Navigation]{\label{fig:a}\includegraphics[width=43mm]{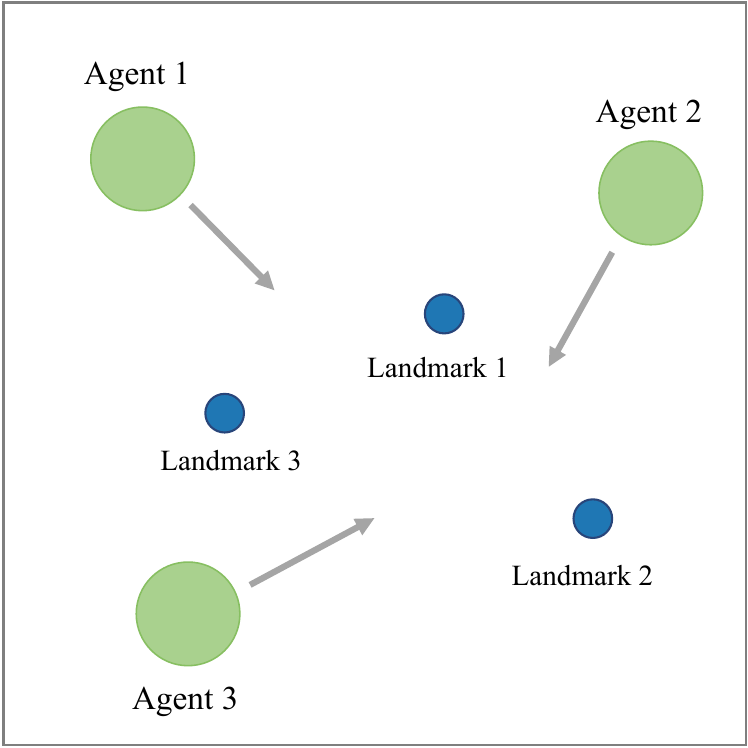}}
\subfigure[Cooperative Treasure Collection]{\label{fig:b}\includegraphics[width=43mm]{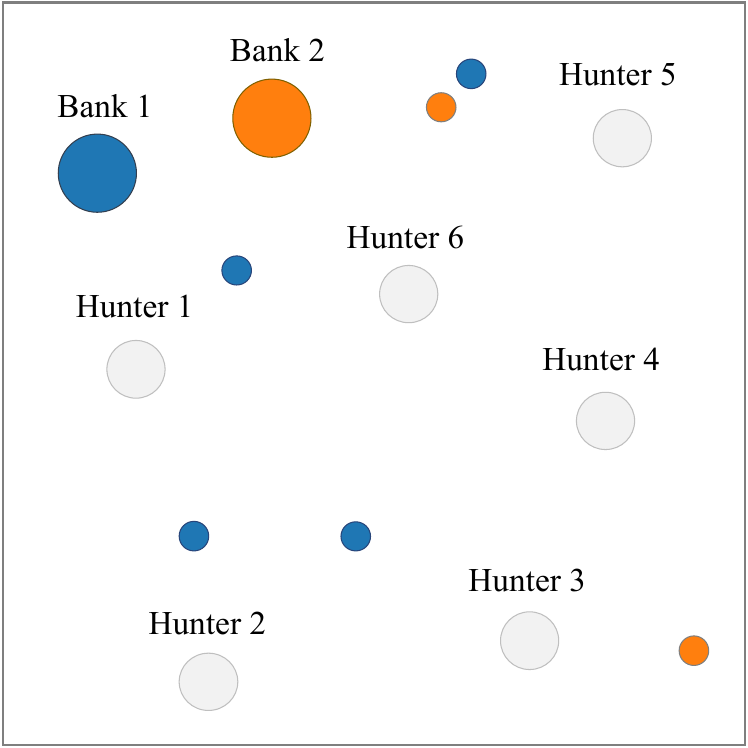}}
\caption[Caption for LOF]{(a) \emph{Cooperative Navigation}. The agents must cover all the landmarks through coordinated movements. While all agents are punished for colliding with each other, only agent 1 is rewarded based on the proximity of any agent to each landmark. 
(b) \emph{Cooperative Treasure Collection}. The hunters need to collect treasures, which are represented by small colored circles, and deposit them into correctly colored banks. The banks are rewarded for the successful collection and depositing of treasure. The hunters do not receive any reward, but are penalized for colliding with each other \cite{mordatch2018emergence}.}
\label{icml-envs}
\end{figure}

\emph{a) Cooperative Navigation:}
$N$ agents must reach a set of $N$ landmarks through coordinated movements. In the original version, agents are collectively rewarded based on the minimum agent distance to each landmark and individually punished when colliding with each other. To make it more challenging for fully decentralized MARL, we modify the task, where only agent $1$ is rewarded; the other agents do not receive rewards but are penalized when collision. Table \ref{coop_nav_obs_rew} summarizes the observations and rewards of the agents.
\vspace{-1em}
\begin{table}[h]
\centering
\caption{The Cooperative Navigation Task.}
\label{coop_nav_obs_rew}
\begin{IEEEeqnarraybox}[\IEEEeqnarraystrutmode\IEEEeqnarraystrutsizeadd{3pt}{3pt}]{x.t.x.t.x.t.x}
\IEEEeqnarrayrulerow[1pt]\\
& Agent && Observations && Reward* & \\
\IEEEeqnarrayrulerow[0.5pt]\IEEEeqnarraystrutsize{1pt}{3pt}\\
& \#1 &&  && $-\min_{i,j}(d(\text{ag}_i, \text{lm}_j))-\mathbbm{1}_{c}$ \IEEEeqnarraystrutsize{1.5pt}{4pt}\\
\IEEEeqnarraymulticol{2}{h}{} && \raisebox{-13pt}[0pt][0pt]{\shortstack{1. position and velocity of itself \\ 2. relative position of landmarks \\ 3. relative position of other agents}} && \IEEEeqnarraymulticol{2}{h}{} \IEEEeqnarraystrutsize{1.5pt}{5pt} \\
& \#2-N &&  && $-\mathbbm{1}_{c}$  & \IEEEeqnarraystrutsize{1.5pt}{4.5pt}\\
\IEEEeqnarrayrulerow[1pt]\IEEEeqnarraystrutsize{2pt}{2pt} \\
&\IEEEeqnarraymulticol{6}{s}{\scriptsize\shortstack{\textsuperscript{*}$d(\text{ag}_i, \text{lm}_j)$ is the Euclidean distance between agent $i$ and landmark $j$.\\\textsuperscript{*}$\mathbbm{1}_{c}$ equals $1$ if the agent collides with another agent and $0$ otherwise.}}
\end{IEEEeqnarraybox}
\end{table}
\vspace{-1em}

\emph{b) Cooperative Treasure Collection:} $N_{h}$ treasure hunters and $N_{b}$ treasure banks search around the environment to gather treasures. The treasures are generated with different colors and re-spawn randomly upon being collected. The hunters are responsible for collecting treasures and depositing them into correctly colored banks. The hunters do not receive any reward from collecting treasures, but they are penalized for colliding with each other. The banks are rewarded based on the function:
\begin{equation}
\label{coop_trea_reward}
\begin{split}
r_{\text{bank}}=&\ 5\big(\textstyle\sum_{j\in\mathcal{H}}\mathbbm{1}_{collect}^{j}+\textstyle\sum_{i\in\mathcal{N}}\mathbbm{1}_{deposit}^{i}\big) \\
&-0.1\min_{j,k}\{d(\text{ht}_j,\text{tr}_{k})|\text{if hunter $j$ is empty}\},
\end{split}
\end{equation}
where $\mathbbm{1}_{collect}^{j}$ is $1$ if a treasure is collected by hunter $j$ and $0$ otherwise; $\mathbbm{1}_{deposit}^{i}$ is $1$ if a treasure is deposited to bank $i$ and $0$ otherwise; $d(\text{ht}_j,\text{tr}_{k})$ is the distance between hunter $j$ and treasure $k$. Table \ref{coop_trea_obs_rew} summarizes the observations and rewards of the agents.

\begin{table}[h]
\centering
\caption{The Cooperative Treasure Collection Task.}
\label{coop_trea_obs_rew}
\begin{IEEEeqnarraybox}[\IEEEeqnarraystrutmode\IEEEeqnarraystrutsizeadd{3pt}{3pt}]{x.t.x.t.x.t.x}
\IEEEeqnarrayrulerow[1pt]\\
& Agent && Observations && Reward* & \\
\IEEEeqnarrayrulerow[0.5pt]\IEEEeqnarraystrutsize{2pt}{3pt}\\
& Banks &&  && Eq. (\ref{coop_trea_reward}) \IEEEeqnarraystrutsize{9pt}{3pt}\\
\IEEEeqnarraymulticol{2}{h}{} && \raisebox{-18pt}[0pt][0pt]{\shortstack{1. position and velocity of itself\ (all agents)\ \ \ \ \\ 2. colors and positions of treasures (all agents) \\ 3. relative positions of other agents (all agents)\\ 4. status of holding a treasure or not (only hunters)}} && \IEEEeqnarraymulticol{2}{h}{} \IEEEeqnarraystrutsize{5pt}{4pt} \\
& Hunters &&  && $-5\times\mathbbm{1}_{c}$  & \IEEEeqnarraystrutsize{9pt}{7pt}\\
\IEEEeqnarrayrulerow[1pt]\IEEEeqnarraystrutsize{2pt}{2pt} \\
&\IEEEeqnarraymulticol{6}{s}{\scriptsize\shortstack{\textsuperscript{*}$\mathbbm{1}_{c}$ is $1$ if the hunter collides with another hunter and $0$ otherwise.}}
\end{IEEEeqnarraybox}
\end{table}
\begin{table}[!t]
\caption{Parameters of MATRPO for the MPE tasks.}
\label{exp_params}
\centering
\begin{tabular}{P{2.2cm}P{1.0cm}P{1.0cm}P{1.0cm}P{1.0cm}}
\toprule
Task & \multicolumn{2}{c}{Cooper. Navigation} & \multicolumn{2}{c}{Cooper. Treas. Collection} \\
\midrule
Number of Agents & $N=3$ & $N=6$ & $N=8$ & $N=12$ \\
Sim. steps per iter. & 10k & 10k & 10k & 10k \\
Stepsize ($\delta/N$). & 0.003 & 0.003 & 0.001 & 0.001 \\
Discount ($\gamma$). & 0.995 & 0.995 & 0.995 & 0.995 \\
GAE Para. ($\lambda$). & 0.95 & 0.95 & 0.95 & 0.95 \\
ADMM iter. & 100 & 150 & 200 & 500 \\
Penalty para. ($\beta$). & 1.0 & 1.0 & 5.0 & 5.0 \\
\bottomrule
\end{tabular}
\end{table}
\subsubsection{Training Performance}
\begin{figure}[t]
\centering
\addtocounter{subfigure}{-1}
\subfigure{\label{fig:d}\includegraphics[width=88mm]{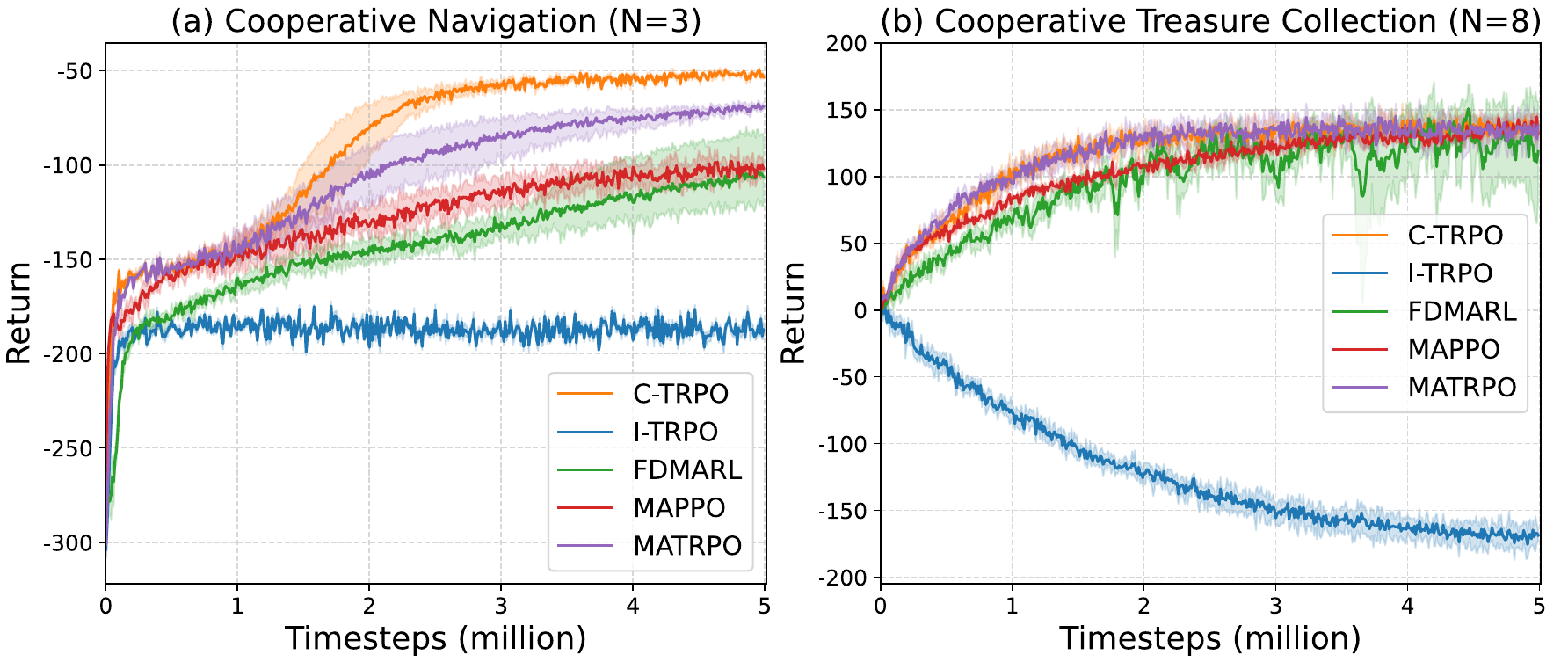}}
\caption[Caption for LOF]{Learning curves over 5 random seeds on the (a) Navigation ($N=3$), and (b) Treasure Collection ($N=8$) tasks.}
\label{learning_curves_1}
\end{figure}

We train the proposed algorithm (MATRPO) and the baseline methods for 5 million timesteps and compared and the episode returns during training. The episode length is 100 timesteps. Other settings of the hyper-parameters are given in Table \ref{exp_params}. For comparison, we further introduce a fully centralized method (C-TRPO), which learns a global policy by reducing the problem to a single-agent RL based on TRPO. The learning curves of MATRPO and the baselines are plotted in Fig.~\ref{learning_curves_1}. As illustrated, MATRPO successfully learned collaborative policies for the networked agents and achieved increased returns. Additionally, MATRPO demonstrated comparable performance with C-TRPO, which nearly finds the global optimum. On the other hand, I-TRPO performed poorly, particularly on the treasure collection task. Since the hunters are penalized for colliding with each other, agents trained by I-TRPO may tend to move far away from each other to avoid collisions, leading to some hunters being far away from the treasures.

Furthermore, MATRPO demonstrates superior performance to FDMARL and similar performance to MAPPO on these tasks. Compared to FDMARL, both MATRPO and MAPPO demonstrate better learning stability as they limit policy updates to a trust region during training. While MAPPO achieves similar final performance to MATRPO, it requires a control center to coordinate training. In contrast, MATRPO is fully decentralized and less dependent on powerful computational resources or a centralized communication network. Although FDMARL is also fully decentralized, its learning process is less stable, and it has a slower learning speed.

\begin{figure}
\centering
\addtocounter{subfigure}{-1}
\subfigure{\label{fig:d}\includegraphics[width=88mm]{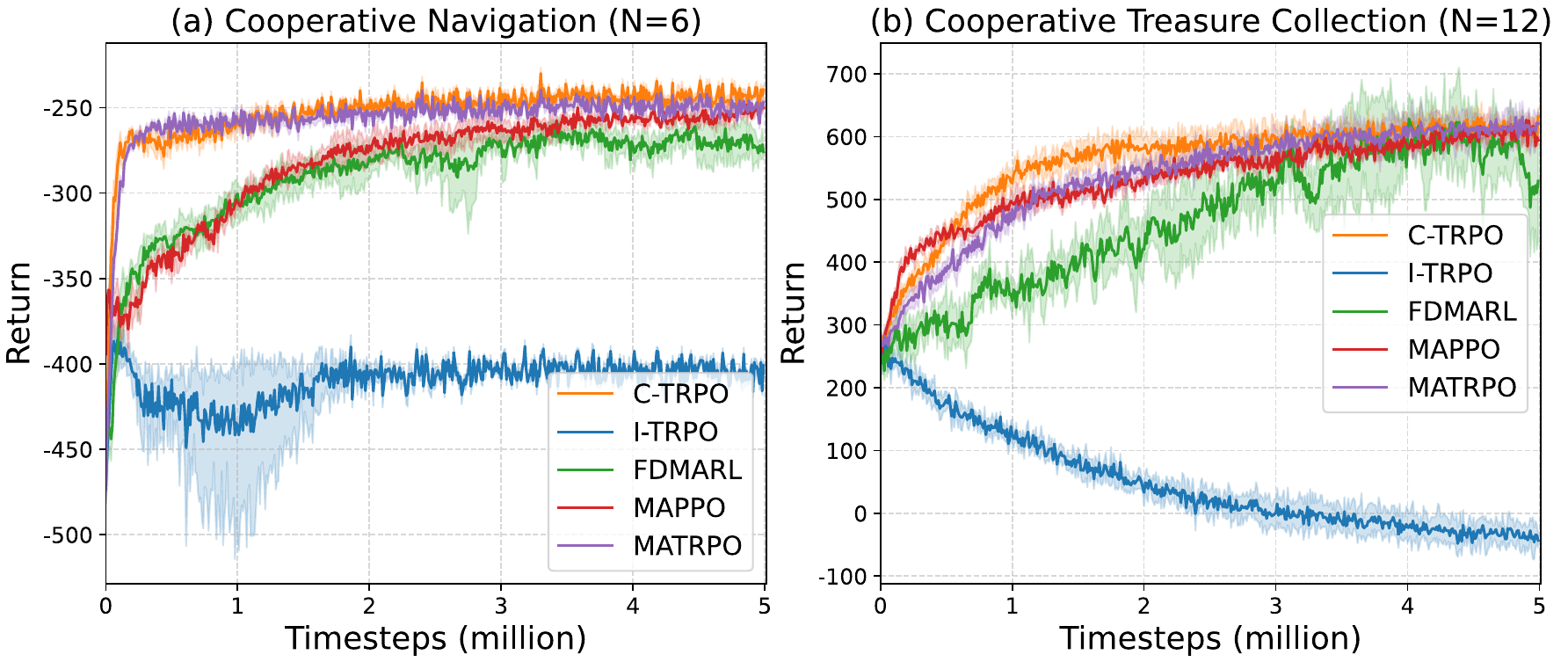}}
\caption[Caption for LOF]{Learning curves over 5 random seeds on the (a) Navigation ($N=6$), and (b) Treasure Collection ($N=12$) tasks.}
\label{learning_curves_2}
\end{figure}

To test the scalability of MATRPO, we conduct additional experiments by increasing the number of agents to $N=6$ and $N=12$ for the two tasks, respectively. The learning curves are plotted in Fig.~\ref{learning_curves_2}. It can be observed that for the tasks with a larger number of agents, MARTPO still achieves comparable or superior performance to the baselines. This result demonstrates the effectiveness of the proposed algorithm when problems are scaled up.

\subsubsection{Impact of Hyperparameters} 
\begin{figure}
\subfigure[Impact of KL stepsizes ($\delta/N$)]{\label{fig:a}\includegraphics[width=87mm]{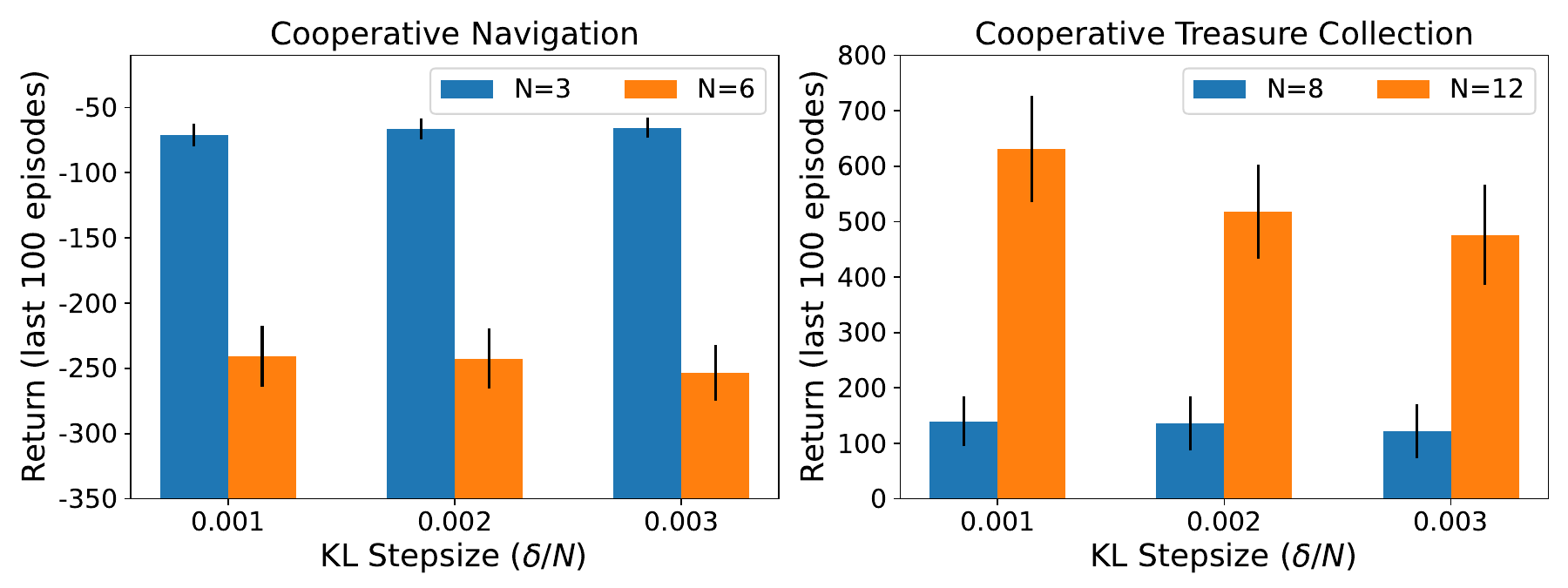}}
\subfigure[Impact of ADMM penalty parameter ($\beta$)]{\label{fig:b}\includegraphics[width=87mm]{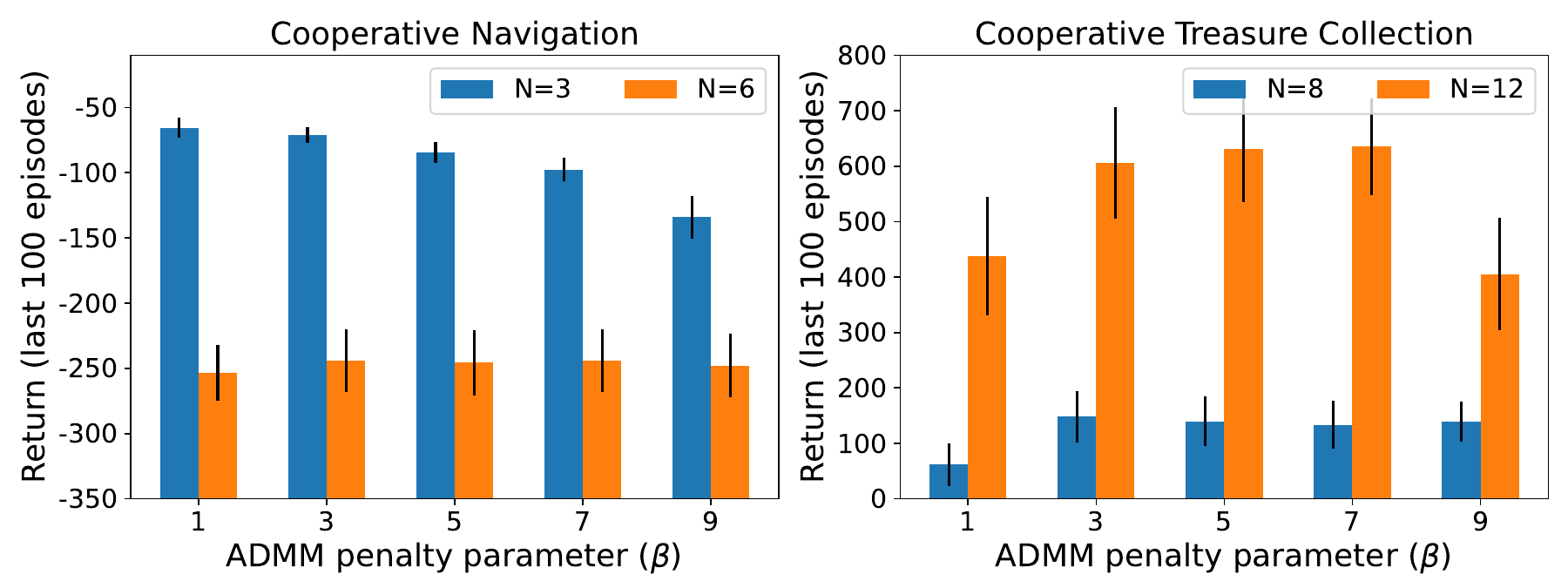}}
\subfigure[Impact of ADMM iterations]{\label{fig:c}\includegraphics[width=87mm]{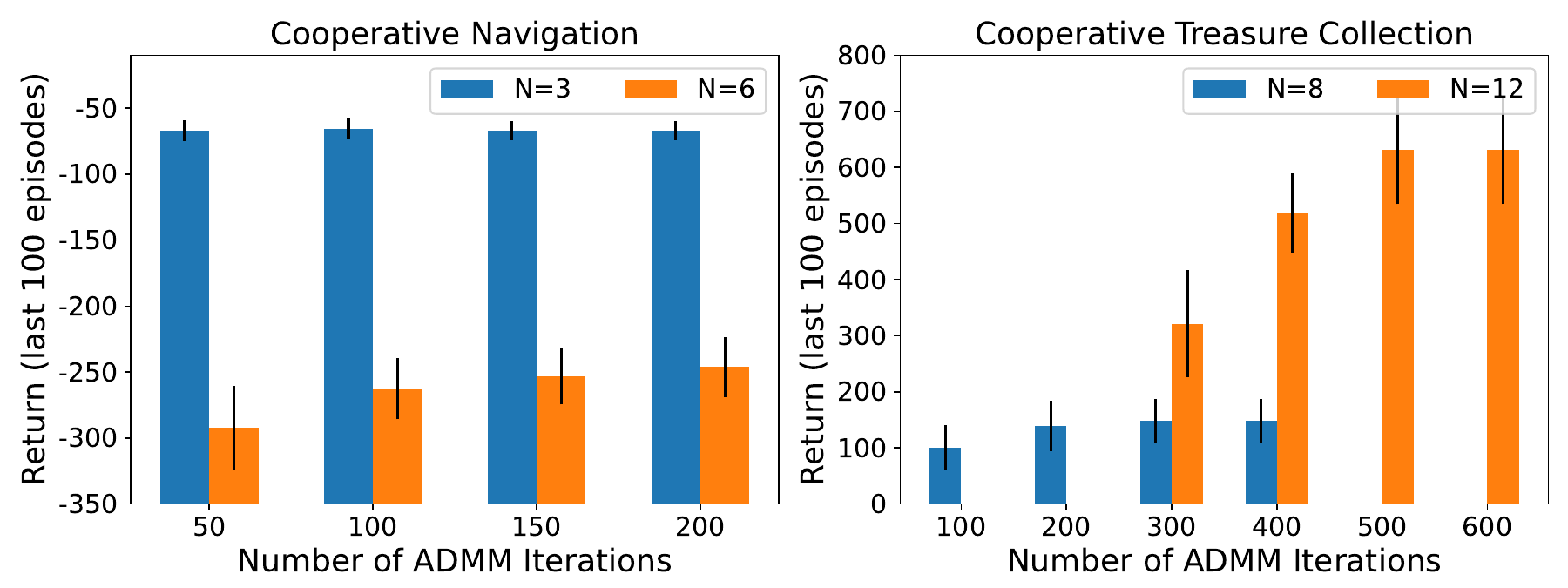}}
\caption[Caption for LOF]{Final performance (returns of the last 100 episodes) of MATRPO using different hyper-parameter settings.}
\label{effect_paras}
\end{figure}

We analyze the influence of three key hyperparameters: 

\emph{a) KL stepsize ($\delta/N$):} The impact of the KL stepsize on MATRPO's final performance is demonstrated in Fig.~\ref{effect_paras}(a). We observe that for the Navigation task with $N=3$ and the Treasure Collection task with $N=8$, KL stepsize has minor effects on the performance. When $N$ increases to $6$ and $12$, respectively, we observe detrimental effects for large KL stepsize, e.g. $\delta/N=0.003$. Note that when $N$ is fixed, a large $\delta/N$ leads to a large $\delta$, which may result in an overly broad trust region for policy search, and thus possible bad updates. This result suggests that small KL stepsizes may be better in problems where there are a large number of agents.

\emph{b) ADMM penalty parameter ($\beta$):} The impact of the ADMM penalty parameter on MATRPO's final performance is demonstrated in Fig.~\ref{effect_paras}(b). We observe that the impact of the penalty parameter $\beta$ varies with the number of agents. For $N=3$, smaller values of $\beta$ can benefit the performance. For larger $N=6, 8, 12$, the performance may be hurt if the penalty parameter is too small ($\beta=1$) or too large ($\beta=9$). Since $\beta$ penalizes disagreement between agents, a small $\beta$ may produce inadequate penalty for a large group of agents to reach consensus within limited iterations; if $\beta$ is too large, the penalty may dominate the objective and have a negative impact on the policy optimization.

\emph{c) Number of ADMM iterations:} The impact of the number of ADMM iterations on MATRPO's final performance is demonstrated in Fig.~\ref{effect_paras}(c). We observe that in nearly all cases, the final performance of MATRPO gets better as the number of ADMM iterations increases; when it reaches a certain point, continuing to increase the number of ADMM iterations does not result in performance improvement. The number of ADMM iterations is a key factor in the distributed ADMM algorithm \cite{Wei2013}, and a large value setting is usually preferred to guarantee consensus. However, once the algorithm converges, doing more ADMM iterations is needless and inefficient in terms of training time and computational resources. 

\begin{figure}[t!]
\centering     
\subfigure[2s3z]{\label{fig:a}\includegraphics[width=43mm]{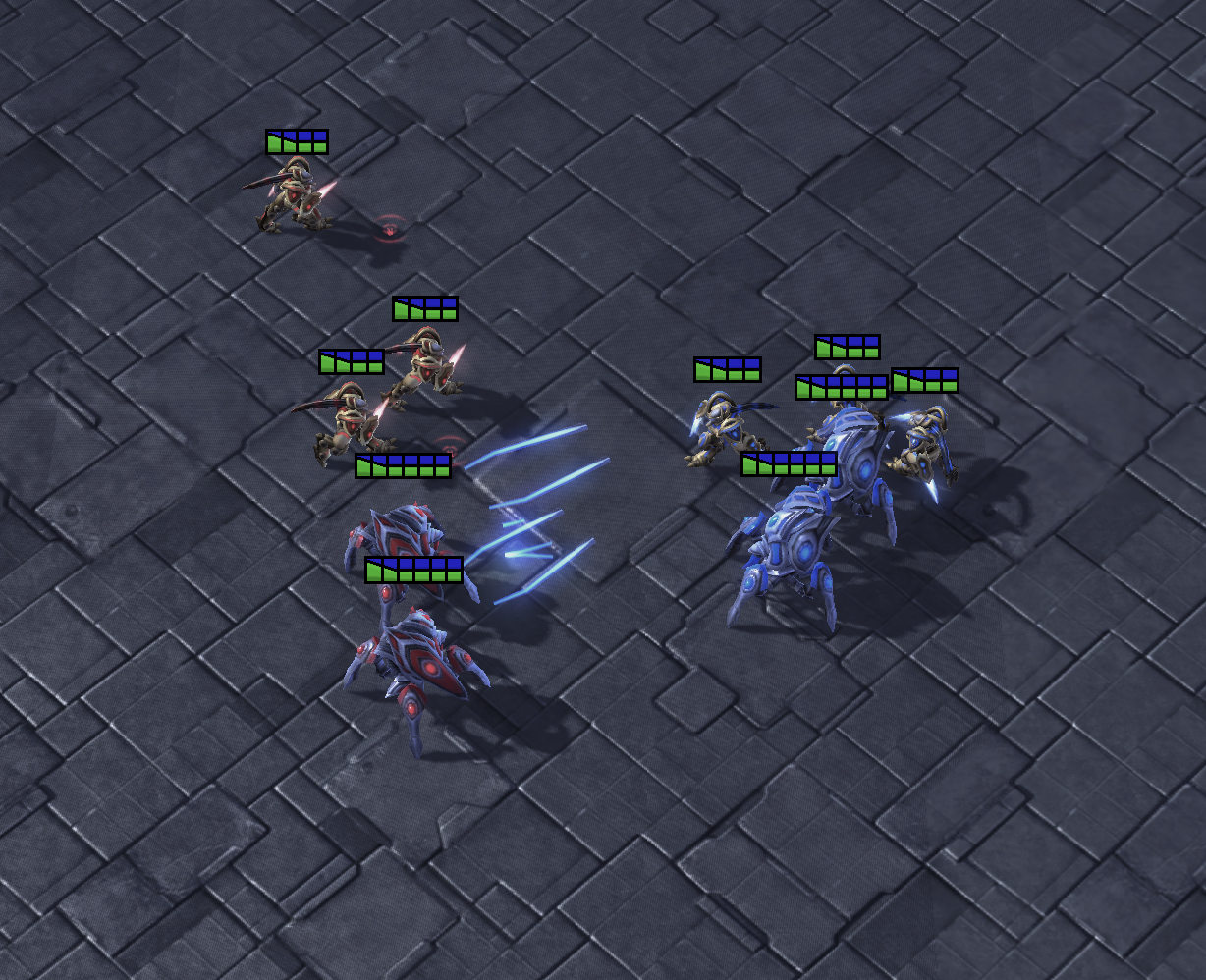}}
\subfigure[3s5z]{\label{fig:b}\includegraphics[width=43mm]{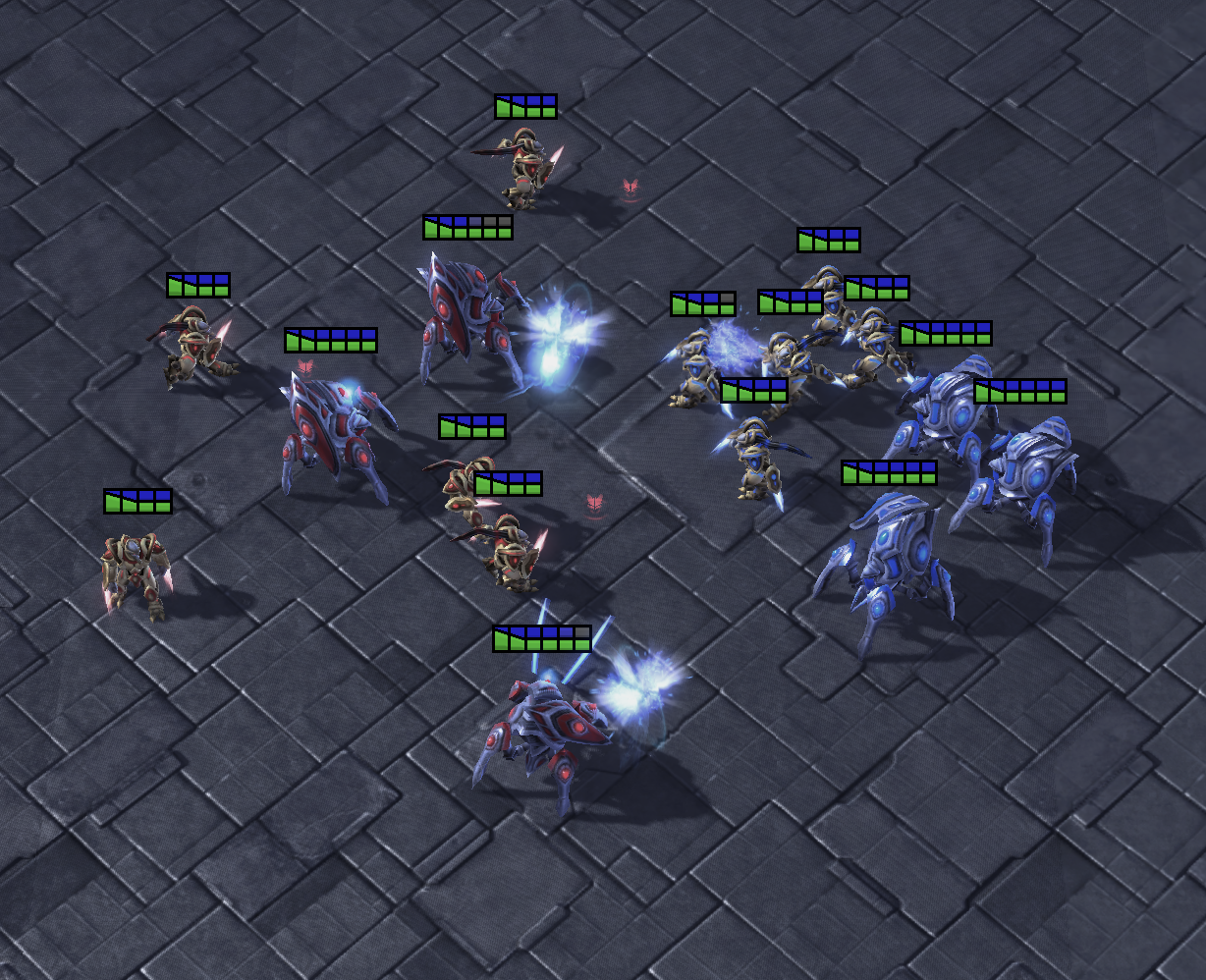}}
\subfigure[10m$\_$vs$\_$11m]{\label{fig:c}\includegraphics[width=43mm]{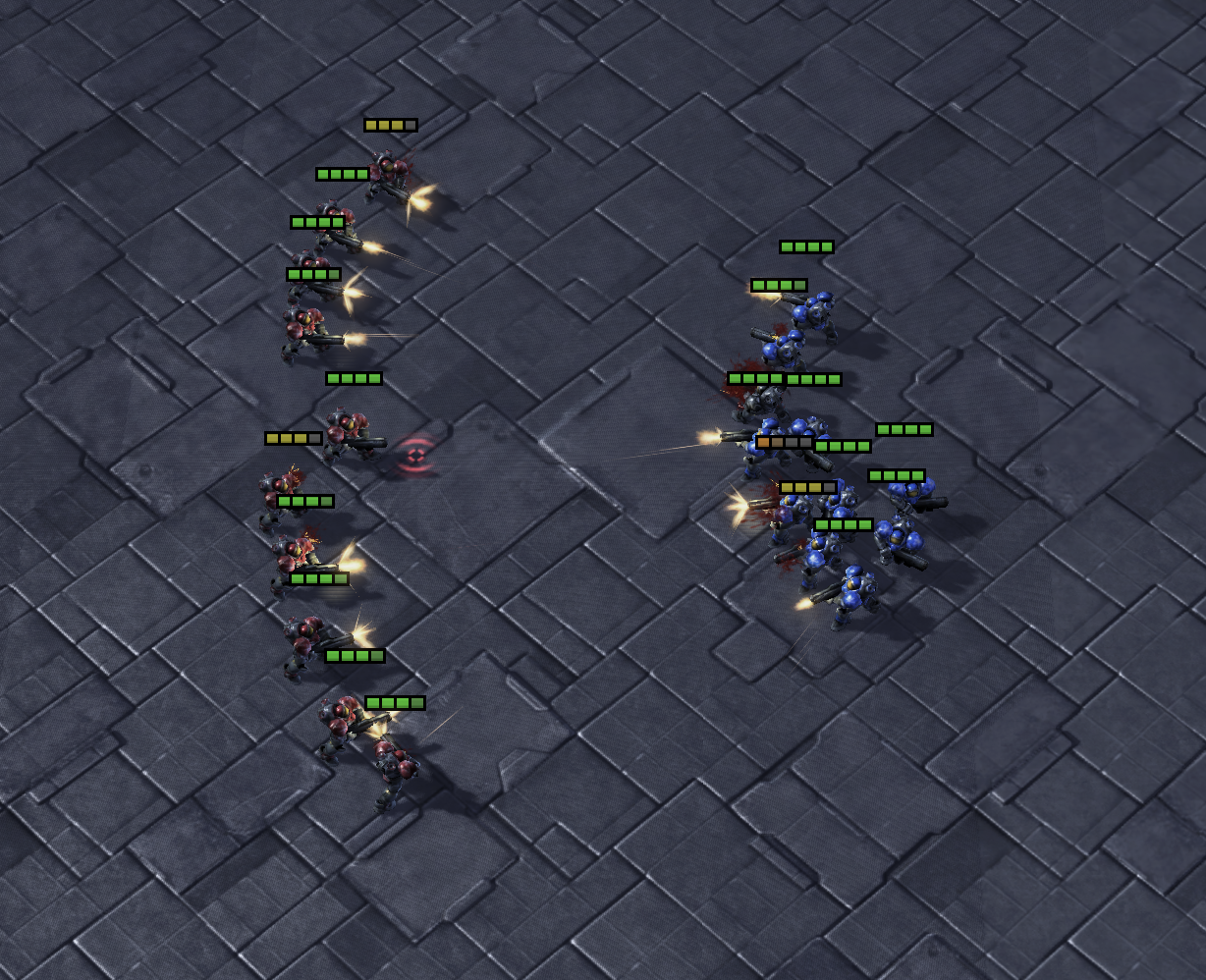}}
\subfigure[MMM2]{\label{fig:c}\includegraphics[width=43mm]{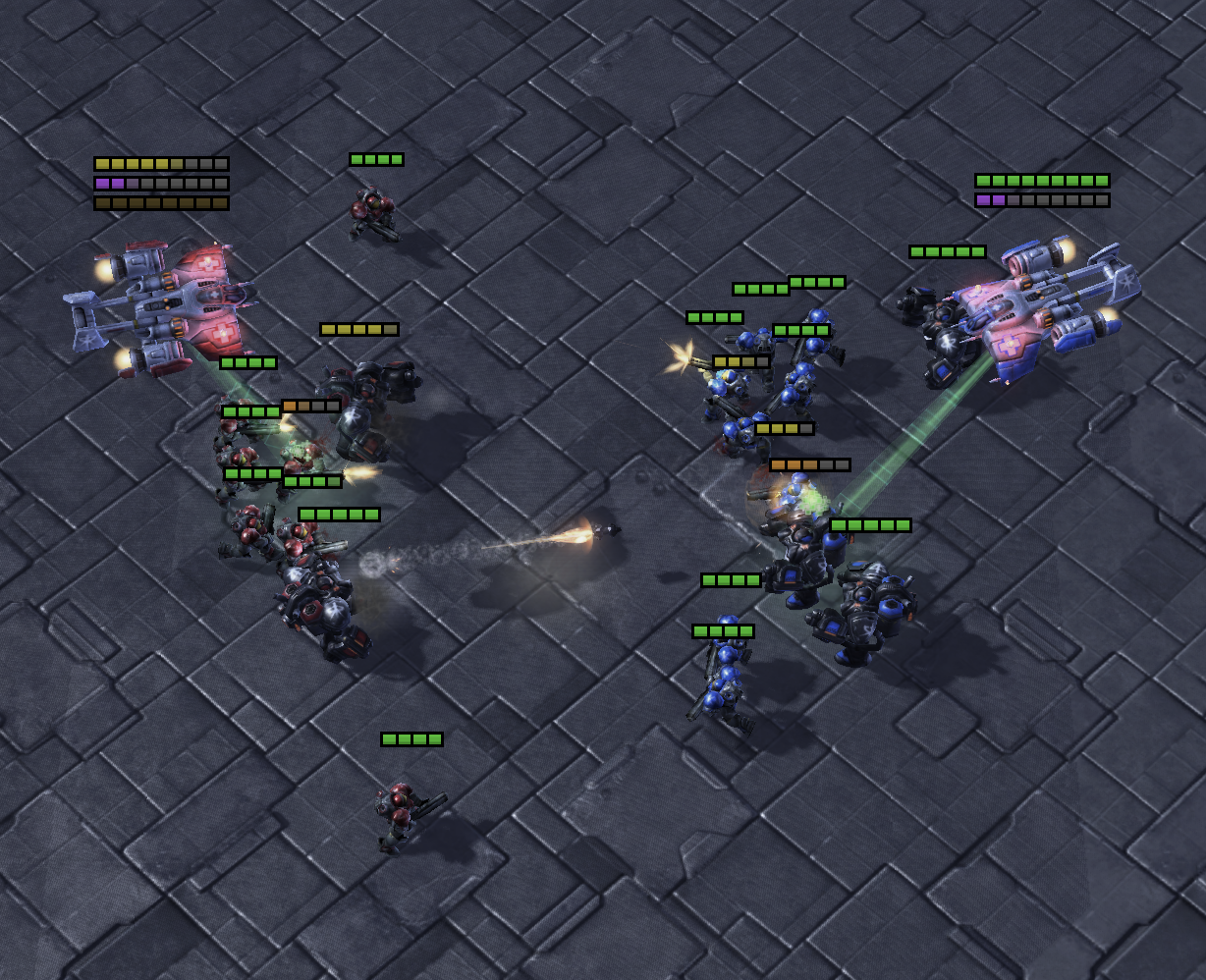}}
\caption[Caption for LOF]{Screenshots of 4 SMAC scenarios: (a) 2s3z, (b) 3s5z, (c) 10m$\_$vs$\_$11m, and (d) MMM2. The red team on the left side is controlled by MARL \cite{samvelyan19smac}.}
\label{scenarions}
\end{figure}
\begin{figure*}[t]
\addtocounter{subfigure}{-1}
\subfigure{\label{fig:a}\includegraphics[width=181mm]{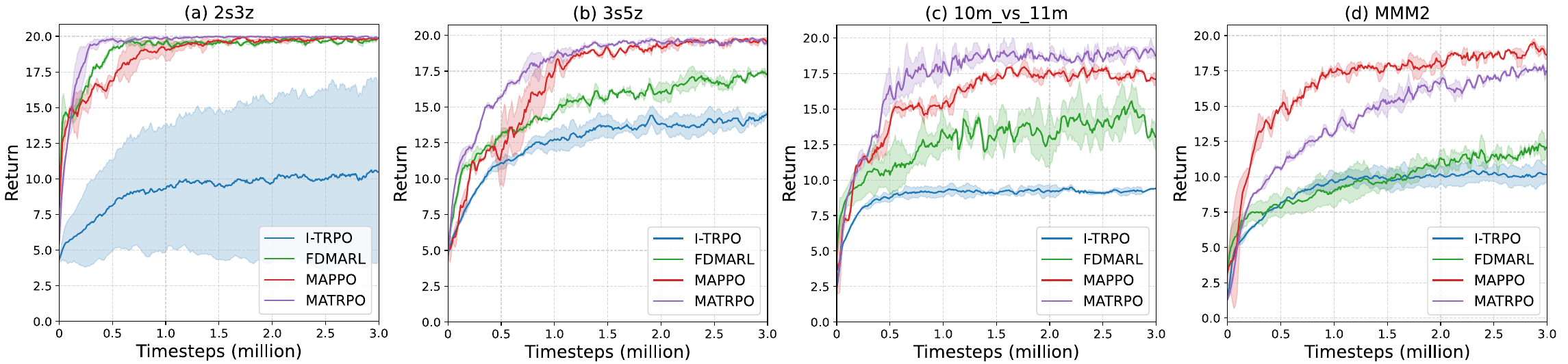}}
\caption[Caption for LOF]{Episode return during the training process over 5 random seeds on the scenarios: (a) 2s3z, (b) 3s5z, (c) 10m$\_$vs$\_$11m, and (d) MMM2.}
\label{smac_return}
\end{figure*}
\begin{figure*}[t]
\centering
\addtocounter{subfigure}{-1}
\subfigure{\label{fig:a}\includegraphics[width=181mm]{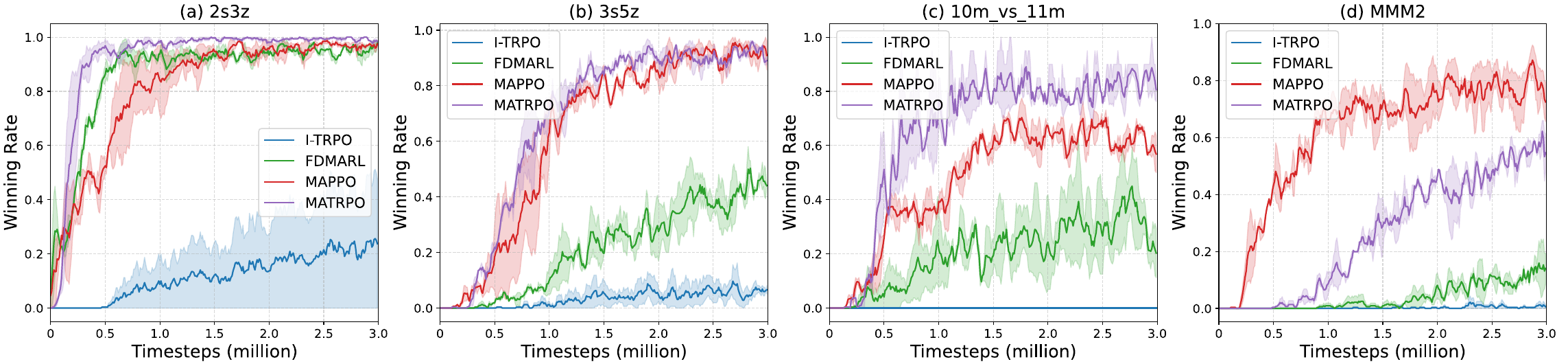}}
\caption[Caption for LOF]{Evaluation winning rate during the training process over 5 random seeds on the scenarios: (a) 2s3z, (b) 3s5z, (c) 10m$\_$vs$\_$11m, and (d) MMM2.}
\label{smac_win_rate}
\end{figure*}
\subsection{StarCraft Multi-Agent Challenge (SMAC)}
\subsubsection{Environment Description} SMAC \cite{samvelyan19smac} is a complex environment for MARL research, focusing on micro-management in StarCraft II. We evaluate MATRPO on four SMAC scenarios (Fig. \ref{scenarions}):

a) \emph{2s3z:} 2 Stalkers \& 3 Zealots are controlled by a group of RL agents, battling an opposing army of 2 Stalkers \& 3 Zealots under the control of the built-in heuristic AI.

b) \emph{3s5z:} A larger version of \emph{2s3z}, where each group has 3 Stalkers \& 5 Zealots. Note that both \emph{2s3z} and \emph{3s5z} scenarios have heterogeneous agents.

c) \emph{10m$\_$vs$\_$11m:} 10 Marines are trained to defeat 11 Marines that are controlled by the heuristic AI. Although the agents are homogeneous, the game is asymmetric and more challenging.

d) \emph{MMM2:} 1 Medivac, 2 Marauders \& 7 Marines are trained to defeat a larger group of enemies with 1 Medivac, 2 Marauders \& 8 Marines. The Medivac is responsible for healing its teammates and other units are responsible for attacking. This task is challenging even for human players since it requires coordinate control of three different types of units. This scenario is both heterogeneous and asymmetric.

At each timestep, agents receive observation information within their sight range (a circular area), including enemy features, ally features, agent movement features, and agent unit features. Attacker units are allowed to take the following actions: move[$direction$], attack[$enemy\_id$], stop and no-op. As healer units, Medivacs take heal[$agent\_id$] actions instead of attack[$enemy\_id$]. We adopt the default setting for a shaped reward in the SMAC environment \cite{samvelyan19smac}, which is calculated from the damage dealt and received by agents, having enemy or allies killed, and winning/losing the battle. To make it challenging for decentralized training, the environment is modified such that only one agent receives the reward signal and the other agents receive 0 reward. This is different than the original setting in SMAC, where all agents receive a share reward. Details on observation features, action space, and reward function are found in the SMAC paper \cite{samvelyan19smac}.
\subsubsection{Training Performance} 
We train the proposed MATRPO algorithm and the baseline methods for 3 million timesteps over 5 random seeds. For each random seed, we run 32 evaluation games after every 10000 timesteps of training and take the evaluation winning rate as an index of the performance. For the hyperparameters of MATRPO, we set $\delta/N=0.001$ and $\beta=5$, and set the number of ADMM iterations to $500$. Other hyper-parameter settings are the same as in the MPE environment.

The episode return and evaluation winning rate during the training process are plotted in Fig.~\ref{smac_return} and Fig.~\ref{smac_win_rate}, respectively. The results indicate that MATRPO outperforms the fully decentralized baselines, I-TRPO and FDMARL, on all scenarios in terms of both episode return and evaluation winning rate. Moreover, on nearly all scenarios, except MMM2, MATRPO achieves comparable performance to the centralized training baseline method, MAPPO. Specifically, on the 2s3z and 3s5z scenarios, MATRPO and MAPPO completely solve the task and achieve over $95$\% winning rate against the built-in heuristic AI. On the 10m$\_$vs$\_$11m scenario, MATRPO achieves over $80$\% winning rate within 1.5 million training timesteps and performs slightly better than MAPPO. On the MMM2 scenario, MATRPO obtains close performance to MAPPO in terms of the final episode return and achieves around $60$\% winning rate. Although MAPPO achieves better performance on this scenario, it uses global observation and shared reward and adopts parameter sharing for agents' policy and value networks during training. On the other hand, MATRPO only uses agents' local observation and reward signals and neighboring communications, thus facilitating a fully decentralized solution to cooperative MARL for networked agents.
\section{Conclusion}
\nocite{conclusion}
In this paper, we have demonstrated how TRPO can be extended to partially observable multi-agent systems through proper decomposition and transformation. This has led to the development of our fully decentralized MARL algorithm, named MATRPO, which optimizes distributed policies for networked agents. Furthermore, we have proven that the distributed policy optimization in the proposed method is equivalent to single-agent TRPO when the local observation of each agent is sufficient. Simulation studies on the MPE tasks and the SMAC battle scenarios have shown that MATRPO is effective at learning collaborative policies in a fully decentralized manner. We have further tested the effectiveness and superiority of the proposed method by comparing it with several baseline methods, including centralized training and decentralized execution, independent learning, and fully-decentralized training. The results have shown that MATRPO outperforms or performs comparably to the baseline methods, demonstrating its potential for scalable and decentralized multi-agent systems.


%

\appendices
\section{}
\label{appendix-A}
\subsection{Equivalent Transformation of Policy Optimization in TRPO}
For a multi-agent system, let $\pi^{i}(a|s)$ be the local policy of agent $i\in\mathcal{N}=\{1,2,\dots,N\}$. When choosing actions, each agent $i$ acts according to the marginal distribution $\pi^{i}(a^{i}|s)$.
\begin{lemma}
\label{prop1}
Assume the local policy has the form $\pi^{i}(a|s)=\pi^{i}(a^{1}|s)\pi^{i}(a^{2}|s)\cdots\pi^{i}(a^{N}|s)$. Then, the TRPO policy update (\ref{decomposed_trpo}) can be equivalently transformed into:
\begin{equation}
\label{matrpo_appendix}
\begin{split}
\max_{\{\pi^{i}\}_{i\in\mathcal{N}}}\ \ &\sum_{i=1}^{N}\underset{\substack{s\sim\rho_{\pi_{old}}\\ a\sim\pi_{old}}}{\mathbb{E}}\left[\frac{\pi^{i}(a|s)}{\pi^{i}_{old}(a|s)}A^{i}_{\pi_{old}}(s,a)\right] \\
s.t.\ \ & \frac{\pi^{1}(a^{i}|s)}{\pi^{1}_{old}(a^{i}|s)}=\cdots=\frac{\pi^{N}(a^{i}|s)}{\pi^{N}_{old}(a^{i}|s)}, \forall i\in\mathcal{N}\\
& \overline{D}_{KL}^{\rho_{\pi_{old}}}(\pi_{old},\pi) \leq \delta.
\end{split}
\raisetag{5.2\normalbaselineskip}
\end{equation}
\end{lemma}
\begin{proof}[Proof] 
We expand the likelihood ratio $\frac{\pi^{i}(a|s)}{\pi^{i}_{old}(a|s)}$ of agent $i$ into
\begin{equation}
\label{likelihood_ratio_expansion}
\frac{\pi^{i}(a|s)}{\pi^{i}_{old}(a|s)}=\frac{\pi^{i}(a^{1}|s)}{\pi^{i}_{old}(a^{1}|s)}\frac{\pi^{i}(a^{2}|s)}{\pi^{i}_{old}(a^{2}|s)}\cdots\frac{\pi^{i}(a^{N}|s)}{\pi^{i}_{old}(a^{N}|s)}.
\end{equation}
When the consensus constraints in (\ref{matrpo_appendix}) are satisfied, we can rewrite the expansion (\ref{likelihood_ratio_expansion}) by replacing $\frac{\pi^{i}(a^{j}|s)}{\pi^{i}_{old}(a^{j}|s)}$ for all $i\neq j$ with $\frac{\pi^{j}(a^{j}|s)}{\pi^{j}_{old}(a^{j}|s)}$ as follows:
\begin{equation}
\label{key_equation}
\frac{\pi^{i}(a|s)}{\pi^{i}_{old}(a|s)}=\frac{\pi^{1}(a^{1}|s)}{\pi^{1}_{old}(a^{1}|s)}\cdots\frac{\pi^{N}(a^{N}|s)}{\pi^{N}_{old}(a^{N}|s)}=\frac{\pi(a|s)}{\pi_{old}(a|s)}
\end{equation}
Substituting (\ref{key_equation}) into the objective of (\ref{matrpo_appendix}), we have
\begin{equation}
\begin{split}
&\ \ \sum_{i=1}^{N}\underset{\substack{s\sim\rho_{\pi_{old}}\\ a\sim\pi_{old}}}{\mathbb{E}}\left[\frac{\pi^{i}(a|s)}{\pi^{i}_{old}(a|s)}A^{i}_{\pi_{old}}(s,a)\right] \\
=&\ \ \underset{\substack{s\sim\rho_{\pi_{old}}\\ a\sim\pi_{old}}}{\mathbb{E}}\left[\frac{\pi(a|s)}{\pi_{old}(a|s)}\sum_{i=1}^{N}A^{i}_{\pi_{old}}(s,a)\right] \\
=&\ \ \underset{\substack{s\sim\rho_{\pi_{old}}\\ a\sim\pi_{old}}}{\mathbb{E}}\left[\frac{\pi(a|s)}{\pi_{old}(a|s)}A_{\pi_{old}}(s,a)\right]
\end{split}
\end{equation}
\end{proof}
\begin{figure}
\centering     
\includegraphics[width=88mm]{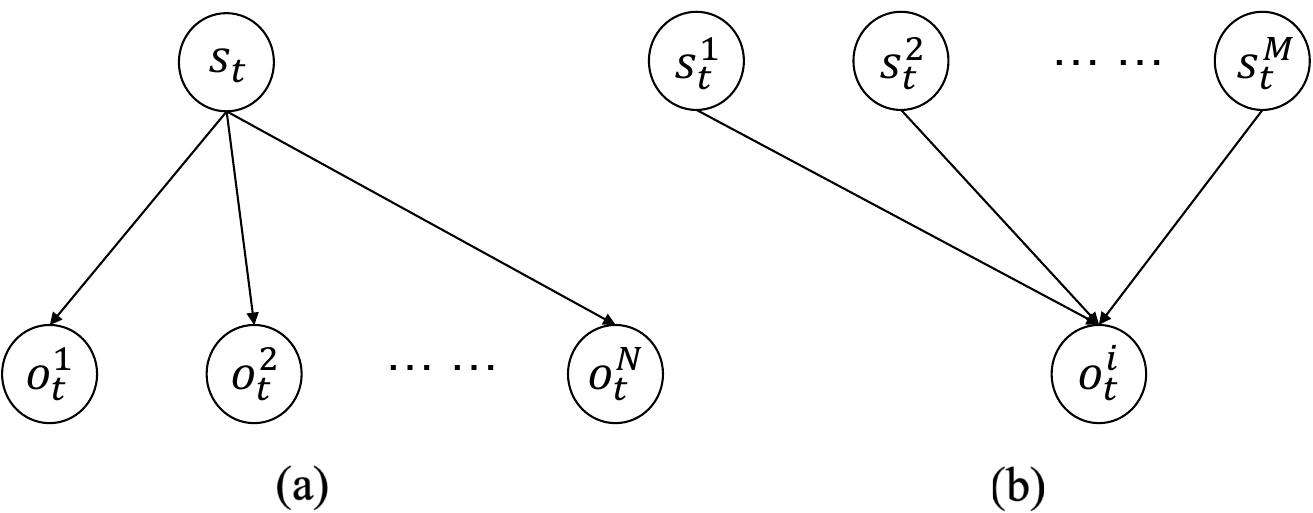}
\caption[Caption for LOF]{In a POMG, (a) agents may have different observations on the same state ). (b) They may also have the same observation on different states.}
\label{po}
\end{figure}

\begin{lemma}
\label{lemma}
If the observation $o^{i},\forall i\in\mathcal{N}$ is sufficient (Definition 1), $\mathbb{E}_{s\sim\rho_{\pi},a\sim\tilde{\pi}}[A^{i}_{\pi}(s,a)]=\mathbb{E}_{o^{i}\sim\rho^{i}_{\pi},a\sim\tilde{\pi}}[A^{i}_{\pi}(o^{i},a)]$ holds.
\end{lemma}
\begin{proof}[Proof]
First note that the discounted state visitation frequencies $\rho_{\pi}$ and the discounted local observation visitation frequencies (defined in Section \ref{preliminaries}-A) have the following relationship:
\begin{equation}
\notag
\label{rhos_to_rhoo}
\begin{aligned}
\sum_{s}\rho_{\pi}(s)P_{o}^{i}(o^{i}|s) 
&= P(o^{i}_{0}=o^{i}|\pi)+\gamma P(o^{i}_{1}=o^{i}|\pi)+\cdots  \\
&= \rho^{i}_{\pi}(o^{i}).
\end{aligned}
\end{equation}

Define $V^{i}_{\pi}(s,o^{i})=\mathbb{E}_{\tau\sim\pi}\left[R^{i}(\tau)\mid s_{0}=s,o^{i}_{0}=o^{i}\right]$ as the local value function of agent $i$ with respect to $(s,o^{i})$. The Bellman equation for $V^{i}_{\pi}(s,o^{i})$ can be expressed by
\begin{equation}
\notag
	V^{i}_{\pi}(s,o^{i})=\sum_{a}\pi(a|o^{i})r^{i}(s,a)+\gamma{\small\sum_{s',o^{i'}}}\mathcal{P}_{\pi}(s',o^{i'}|s,o^{i})V^{i}_{\pi}(s',o^{i'})
\end{equation}
where $\mathcal{P}_{\pi}(s',o^{i'}|s,o^{i})=\sum_{a}\pi^{i}(a|o^{i})\mathcal{P}(s'|s,a)\mathcal{P}_{o}^{i}(o^{i'}|s')$ is the transition probability. Then, the local value functions $V^{i}_{\pi}(s)$ and $V^{i}_{\pi}(o^{i})$ can be expressed respectively as:
\begin{equation}
\begin{split}
	&V^{i}_{\pi}(s)=\sum_{o^{i}}\mathcal{P}^{i}_{o}(o^{i}|s)V^{i}_{\pi}(s,o^{i})= \\
	&\sum_{o^{i}}\mathcal{P}^{i}_{o}(o^{i}|s)\sum_{a}\pi(a|o^{i})r^{i}(s,a)+\gamma{\small\sum_{s',o^{i'}}}\mathcal{P}_{\pi}(s',o^{i'}|s)V^{i}_{\pi}(s',o^{i'}),
\end{split}
\raisetag{4.0\normalbaselineskip}
\end{equation}
\begin{equation}
\begin{split}
	&V^{i}_{\pi}(o^{i})=\sum_{s}\mathcal{P}_{s}(s|o^{i})V^{i}_{\pi}(s,o^{i})= \\
	&\sum_{s}\mathcal{P}_{s}(s|o^{i})\sum_{a}\pi(a|o^{i})r^{i}(s,a)+\gamma{\small\sum_{s',o^{i'}}}\mathcal{P}_{\pi}(s',o^{i'}|o^{i})V^{i}_{\pi}(s',o^{i'}).
\end{split}
\raisetag{4.0\normalbaselineskip}
\end{equation}
Based on the sufficient information assumptions on local observations: 
\textbf{(1)} $\mathbb{E}_{o^{i}\sim\mathcal{P}^{i}_{o},a\sim\pi}[r^{i}(s,a)|s]=\mathbb{E}_{s\sim\mathcal{P}_{s},a\sim\pi}[r^{i}(s,a)|o^{i}]$; \textbf{(2)} $\mathcal{P}_{\pi}(s',o^{i'}|s)=\mathcal{P}_{\pi}(s',o^{i'}|o^{i})$, we can easily prove that 
\begin{equation}
	V^{i}_{\pi}(s)=V^{i}_{\pi}(o^{i}).
\end{equation}
Letting $\mathcal{P}^{i}:\mathcal{O}^{i}\times\mathcal{A}^{1}\times\cdots\times\mathcal{A}^{N}\times\mathcal{O}^{i}\mapsto\left[0,1\right]$ denote the transition distribution of the observations of agent $i$, then the local advantage function can be rewritten by
\begin{equation}
\label{adv_decomposition}
\begin{aligned}
	&A^{i}_{\pi}(s,a)\\
	 =&\ r^{i}(s,a)-V(s)+\gamma\mathbb{E}_{s'\sim\mathcal{P}(\cdot|s,a)}\left[V(s')\right] \\
	=&\ r^{i}(s,a)-V(o^{i})+\gamma\mathbb{E}_{s'\sim\mathcal{P}(\cdot|s,a)}[\mathbb{E}_{o^{i'}\sim\mathcal{P}^{i}(\cdot|o^{i},a)}[V(s')]] \\
	=&\ r^{i}(s,a)-V(o^{i})+\gamma\mathbb{E}_{s'\sim\mathcal{P}(\cdot|s,a)}[\mathbb{E}_{o^{i'}\sim\mathcal{P}^{i}(\cdot|o^{i},a)}[V(o^{i'})]] \\
	=&\ r^{i}(s,a)-V(o^{i})+\gamma\mathbb{E}_{o^{i'}\sim\mathcal{P}^{i}(\cdot|o^{i},a)}[V(o^{i'})] \\
\end{aligned}
\end{equation}
Then, based on Eqs. (\ref{rhos_to_rhoo}) and (\ref{adv_decomposition}), we have
\begin{equation}
\begin{split}
	&\mathbb{E}_{s\sim\rho_{\pi},a\sim\tilde{\pi}}[A^{i}_{\pi}(s,a)]=\sum_{s}\rho_{\pi}(s)\sum_{a}\tilde{\pi}(a|s)A^{i}_{\pi}(s,a) \\
	=&\ \sum_{s}\rho_{\pi}(s)\sum_{a}\Big(\sum_{o^{i}}\tilde{\pi}(a|o^{i})\mathcal{P}_{o}^{i}(o^{i}|s)\Big)A^{i}_{\pi}(s,a) \\	
	=&\ \sum_{o^{i}}\sum_{a}\tilde{\pi}(a|o^{i})\sum_{s}\rho_{\pi}(s)\mathcal{P}^{i}_{o}(o^{i}|s)r^{i}(s,a)\ \ (\text{use (\ref{rhos_to_rhoo}),\ (\ref{adv_decomposition})}) \\
	&\ -\sum_{o^{i}}\sum_{a}\tilde{\pi}(a|o^{i})\rho^{i}_{\pi}(o^{i})\Big[V(o^{i}) -\gamma\mathbb{E}_{o^{i'}\sim\mathcal{P}^{i}(\cdot|o^{i},a)}[V(o^{i'})]\Big] \\
	=&\ \sum_{s}\rho_{\pi}(s)\sum_{o^{i}}\mathcal{P}^{i}_{o}(o^{i}|s)\sum_{a}\tilde{\pi}(a|o^{i})r^{i}(s,a) \\
	&\ -\sum_{o^{i}}\rho^{i}_{\pi}(o^{i})\sum_{a}\tilde{\pi}(a|o^{i})\Big[V(o^{i}) -\gamma\mathbb{E}_{o^{i'}\sim\mathcal{P}^{i}(\cdot|o^{i},a)}[V(o^{i'})]\Big] \\
	=&\ \sum_{o^{i}}\rho_{\pi}^{i}(o^{i})\sum_{s}\mathcal{P}_{s}(s|o^{i})\sum_{a}\tilde{\pi}(a|o^{i})r^{i}(s,a) \\
	&\ -\sum_{o^{i}}\rho^{i}_{\pi}(o^{i})\sum_{a}\tilde{\pi}(a|o^{i})\Big[V(o^{i}) -\gamma\mathbb{E}_{o^{i'}\sim\mathcal{P}^{i}(\cdot|o^{i},a)}[V(o^{i'})]\Big] \\
	=&\ \mathbb{E}_{o^{i}\sim\rho^{i}_{\pi},a\sim\tilde{\pi}}[A^{i}_{\pi}(o^{i},a)].
\end{split}
\raisetag{1.0\normalbaselineskip}
\end{equation}
where $J^{i}(\pi)$ is the expected discounted rewards of agent $i$.
\end{proof}
\begin{manualtheoreminner}
\label{main_result}
Assuming that the local observations $o^{i},\forall i\in\mathcal{N}$ are sufficient, then the distributed optimization problem (\ref{pomatrpo}) is equivalent to the problem (\ref{decomposed_trpo}) with a stricter trust region, i.e., $\overline{D}_{KL}^{\rho_{\pi_{old}}}(\pi_{old},\pi) \leq \Delta$, where $\Delta\leq\delta$.
\end{manualtheoreminner}
\begin{proof}[Proof]
First, we show that the objective of problem (\ref{pomatrpo}) is equivalent to that of (\ref{decomposed_trpo}) if the local observation $o^{i}$ is sufficient.

Note that the objective of (\ref{decomposed_trpo}) can be rewritten as
\begin{equation}
\label{approx_cen_obj}
\begin{split}
	& \underset{\substack{s\sim\rho_{\pi_{old}}\\ a\sim\pi_{old}}}{\mathbb{E}}\left[\frac{\pi(a|s)}{\pi_{old}(a|s)}\left(A^{1}_{\pi_{old}}(s,a)+\cdots+A^{N}_{\pi_{old}}(s,a)\right)\right] \\
    =& \sum_{i=1}^{N}\mathbb{E}_{o^{i}\sim\rho_{\pi_{old}^{i}},a\sim\pi}\left[A^{i}_{\pi_{old}}(o^{i},a)\right]\ \ \ \text{(use \textbf{Lemma \ref{lemma})}} \\
    =& \sum_{i=1}^{N}\underset{\substack{o^{i}\sim\rho_{\pi_{old}}^{i}\\ a\sim\pi_{old}}}{\mathbb{E}}\left[\frac{\pi(a|o^{i})}{\pi_{old}(a|o^{i})}A^{i}_{\pi_{old}}(o^{i},a)\right].
\end{split}
\end{equation}

Using Lemma \ref{prop1}, Eq. (\ref{approx_cen_obj}) can be equivalently transformed into the following objective with consensus constraints:
\begin{equation}
\label{approx_obj}
\begin{aligned}
\text{maximize}\ \ &\sum_{i=1}^{N}\underset{\substack{o^{i}\sim\rho_{\pi_{old}}^{i}\\ a\sim\pi_{old}}}{\mathbb{E}}\left[\frac{\pi^{i}(a|o^{i})}{\pi^{i}_{old}(a|o^{i})}A^{i}_{\pi_{old}}(o^{i},a)\right] \\
\text{subject to}\ \ & \frac{\pi^{1}(a^{i}|o^{1})}{\pi^{1}_{old}(a^{i}|o^{1})}=\cdots=\frac{\pi^{N}(a^{i}|o^{N})}{\pi^{N}_{old}(a^{i}|o^{N})}, \forall i\in\mathcal{N}
\end{aligned}
\end{equation}

Next, we show that the KL constraint over the joint policy in problem (\ref{decomposed_trpo}) will be enforced within a stricter trust region when the individual KL constraints over the local policy are satisfied in the distributed optimization problem (\ref{pomatrpo}). Note that
\begin{equation}
\begin{split}
	&D_{KL}(\pi^{i}_{old}(\cdot|s)\parallel\pi^{i}(\cdot|s) = \sum_{a^{i}}\pi^{i}_{old}(a^{i}|s)\log\frac{\pi^{i}_{old}(a^{i}|s)}{\pi^{i}(a^{i}|s)} \\
	=\ &\sum_{a^{i}}\sum_{o^{i}}\mathcal{P}_{o}^{i}(o^{i}|s)\pi^{i}_{old}(a^{i}|o^{i})\log\frac{\sum_{o^{i}}\mathcal{P}_{o}^{i}(o^{i}|s)\pi^{i}_{old}(a^{i}|o^{i})}{\sum_{o^{i}}\mathcal{P}_{o}^{i}(o^{i}|s)\pi^{i}(a^{i}|o^{i})} \\
	\leq\ &\sum_{a^{i}}\sum_{o^{i}}\mathcal{P}_{o}^{i}(o^{i}|s)\pi^{i}_{old}(a^{i}|o^{i})\log\frac{\mathcal{P}_{o}^{i}(o^{i}|s)\pi^{i}_{old}(a^{i}|o^{i})}{\mathcal{P}_{o}^{i}(o^{i}|s)\pi^{i}(a^{i}|o^{i})} \\
	=\ &\sum_{o^{i}}\mathcal{P}_{o}^{i}(o^{i}|s)\sum_{a^{i}}\pi^{i}_{old}(a^{i}|o^{i})\log\frac{\pi^{i}_{old}(a^{i}|o^{i})}{\pi^{i}(a^{i}|o^{i})} \\
	=\ &\sum_{o^{i}}\mathcal{P}_{o}^{i}(o^{i}|s)D_{KL}(\pi^{i}_{old}(\cdot|o^{i})\parallel\pi^{i}(\cdot|o^{i}),
\end{split}
\raisetag{1.7\normalbaselineskip}
\end{equation}
where the inequality is derived from the logarithmic property 
\begin{equation}
	b\log\frac{b}{c}\leq\sum_{i=1}^{n}b_{i}\log\frac{b_{i}}{c_{i}},
\end{equation}
and $b_{i},c_{i}$ are nonnegative numbers, $b$ is the sum of $b_{i}$s and $c$ is the sum of $c_{i}$s. Based on the additive and non-negative properties of KL-Divergence, we have
\begin{equation}
\begin{split}
&\overline{D}_{KL}^{\rho_{\pi_{old}}}(\pi_{old},\pi) = \mathbb{E}_{s\sim\rho_{\pi_{old}}}\left[D_{KL}(\pi_{old}(\cdot|s)\parallel\pi(\cdot|s))\right]) \\
=\ &\sum_{i=1}^{N}\mathbb{E}_{s\sim\rho_{\pi_{old}}}[D_{KL}(\pi^{i}_{old}(a^{i}|s)\parallel\pi^{i}(a^{i}|s)] \\
\leq\ &\sum_{i=1}^{N}\sum_{s}\rho_{\pi_{old}}(s)\Big[\sum_{o^{i}}\mathcal{P}_{o}^{i}(o^{i}|s)D_{KL}(\pi^{i}_{old}(\cdot|o^{i})\parallel\pi^{i}(\cdot|o^{i})\Big] \\
=\ &\sum_{i=1}^{N}\mathbb{E}_{o^{i}\sim\rho^{i}_{\pi_{old}}}[D_{KL}(\pi^{i}_{old}(\cdot|o^{i})\parallel\pi^{i}(\cdot|o^{i})] \\
=\ &\sum_{i=1}^{N}\overline{D}_{KL}^{\rho^{i}_{\pi_{old}}}(\pi^{i}_{old},\pi^{i}). \\
\end{split}
\raisetag{1.7\normalbaselineskip}
\end{equation}
Using the KL constraints on the local policy in (\ref{pomatrpo}), we have
\begin{equation}
	\overline{D}_{KL}^{\rho_{\pi_{old}}}(\pi_{old},\pi)\leq\sum_{i=1}^{N}\overline{D}_{KL}^{\rho^{i}_{\pi_{old}}}(\pi^{i}_{old},\pi^{i})\leq\sum_{i=1}^{N}\delta/N=\delta.
\end{equation}
Denoting $\Delta=\sum_{i=1}^{N}\overline{D}_{KL}^{\rho^{i}_{\pi_{old}}}(\pi^{i}_{old},\pi^{i})$, the result follows.
\end{proof}
\subsection{Analytical Solution to ADMM Updates}
\label{appendix-B}
Consider the augmented Lagrangian $\mathcal{L}_{\beta}(\bm{\theta},\bm{z},\bm{y})$ in (\ref{sample_augmented_lagrange}), and the asynchronous ADMM updates for:
\begin{equation}
\label{asyn_admm_appendix}
\begin{split}
&\theta^{q(k+1)} :=\argmin_{\theta^{q}\in\Theta^{q}}\mathcal{L}_{\beta}(\bm{\theta},\bm{z}^{(k)},\bm{y}^{(k)}), \\
&z^{qn(k+1)}_{e} :=\argmin_{z^{qn}_{e}\in Z_{e}}\mathcal{L}_{\beta}(\bm{\theta}^{(k+1)},\bm{z},\bm{y}^{(k)}), \\
&y^{qn(k+1)}_{e}:=y^{qn(k)}_{e}+\beta(C_{e}^{q}J^{qn}(\theta^{q(k+1)}-\theta^{q}_{old})-z_{e}^{qn(k+1)}).
\end{split}
\raisetag{3.5\normalbaselineskip}
\end{equation}
where $\Theta^{q}=\{\theta\vert\frac{1}{2}(\theta-\theta^{q}_{old})^{T}\overline{H}^{q}(\theta-\theta^{q}_{old})\leq\delta/N\}$ and $Z_{e}^{n}$ $=\{z_{e}^{in},z_{e}^{jn}\vert z_{e}^{in}+z_{e}^{jn}=0, q=(i,j)\in\mathcal{N}(e)\}$. Next, we give the analytical solutions to the ADMM updates.
\begin{prop}
If there exists at least one strictly feasible point in the feasible sets $\Theta^{i}$ and $Z_{e}^{n}$, the optimal solution of (\ref{asyn_admm_appendix}) is:
\begin{equation}
\label{log_appr_asy_admm_updates_appendix}
\begin{split}
\theta^{q(k+1)}&:=\ \theta^{q}_{old}+\sqrt{\frac{2\delta/N}{V^{T}{\overline{H}^{q}}^{\ -1}V}}{\overline{H}^{q}}^{\ -1}V, \\
z_{e}^{qn(k+1)}&:=\ \frac{1}{\beta}(y_{e}^{qn(k)}-\nu_{e}^{n})+C_{e}^{q}J^{qn}(\theta^{q(k+1)}-\theta^{q}_{old}), \\
y_{e}^{qn(k+1)}&:=\ \nu_{e}^{n}
\end{split}
\raisetag{1.1\normalbaselineskip}
\end{equation}
when $M\rightarrow\infty$, where
\begin{equation}
\notag
V=\frac{1}{M}\sum_{n=1}^{N}{J^{qn}}^{\text{T}}(A^{q}-\sum_{e\in\mathcal{E}(q)}C_{e}^{q}y_{e}^{qn(k)}+\beta\sum_{e\in\mathcal{E}(q)} C_{e}^{q}z_{e}^{qn(k)}),
\end{equation}
\begin{equation}
\begin{split}
\notag
\nu_{e}^{n}=&\frac{1}{2}\sum_{q\in\mathcal{N}(e)}\left[y_{e}^{qn(k)}+\beta C_{e}^{q}J^{qn}(\theta^{q(k+1)}-\theta^{q}_{old})\right].
\end{split}
\end{equation}
and $\mathcal{E}(q)$ is the set of the communication links connectting to agent $q$.
\end{prop}

\begin{proof}[Proof]
Define $x^{q}=\theta^{q}-\theta^{q}_{old}$ and $x^{q}\in X^{q}=\{x\vert\frac{1}{2}x^{\text{T}}\overline{H}^{q}x\leq\delta/N\}$. The Lagrangian (\ref{sample_augmented_lagrange}) can be simplified as
\begin{equation}
\label{simplied_lagrange_appendix}
\begin{split}
\mathcal{L}_{\beta}(\bm{x},\bm{z},\bm{y})=&-\frac{1}{M}\sum_{i=1}^{N}\sum_{n=1}^{N}{A^{i}}^{T} J^{in}x^{i} \\
&+\frac{1}{M}\sum_{e=1}^{L}\sum_{q\in\mathcal{N}(e)}\sum_{n=1}^{N}{y_{e}^{qn}}^{T}(C_{e}^{q}J^{qn}x^{q}-z_{e}^{qn}) \\
&+\frac{1}{M}\sum_{e=1}^{L}\sum_{q\in\mathcal{N}(e)}\sum_{n=1}^{N}\frac{\beta}{2}\parallel C_{e}^{q}J^{q}x^{q}-z_{e}^{qn}\parallel_{2}^{2},
\end{split}
\raisetag{8.0\normalbaselineskip}
\end{equation}

(1) We first consider the update of $\theta^{q}$, which is transformed into the following problem
\begin{equation}
\min_{x^{q}\in X^{q}}\mathcal{L}_{\beta}(\bm{x},\bm{z},\bm{y}),\ q\in\mathcal{N}(e).
\end{equation}
This is a convex optimization with quadratic constraint. According to strong duality theory, the optimal value $p^*$ satisfies
\begin{equation}
\label{lagrange_x}
\begin{split}
p^{*}=&\min_{x^{q}}\max_{\lambda^{q}>0}\ \mathcal{L}_{\beta}(\bm{x},\bm{z},\bm{y})+\lambda^{q}(\frac{1}{2}x^{\text{T}}\overline{H}^{q}x - \delta/N) \\
=&\max_{\lambda^{q}>0}\min_{x^{q}}\ \mathcal{L}_{\beta}(\bm{x},\bm{z},\bm{y})+\lambda^{q}(\frac{1}{2}x^{\text{T}}\overline{H}^{q}x - \delta/N) \\
=&\max_{\lambda^{q}>0}\min_{x^{q}}\ f(x^{q}, \lambda^{q})
\end{split}
\end{equation}
Based on the first order condition, the optimal solution ${x^{q}}^{*}$ should satisfy $\partial f(x^{q}, \lambda^{q}) / \partial x^{q}=0$, i.e.
\begin{equation}
\label{first_ordder_cond_x}
\begin{split}
&\left(\lambda^{q}\overline{H}^{q}+\frac{|\mathcal{E}(q)|\beta}{M}\sum_{n=1}^{N}{J^{qn}}^{T}J^{qn}\right)x^{q}\ - \\
&\frac{1}{M} \sum_{n=1}^{N}{J^{qn}}^{T}\left(A^{q}-\sum_{e\in\mathcal{E}(q)}C_{e}^{q}y_{e}^{qn}+\beta\sum_{e\in\mathcal{E}(q)}C_{e}^{q}z_{e}^{qn(k)}\right) \\
=&\ 0
\end{split}
\end{equation}
where $|\mathcal{E}(q)|$ is the cardinality of $\mathcal{E}(q)$. Solving (\ref{first_ordder_cond_x}), we get
\begin{equation}
{x^{q}}^{*} = \left({\lambda^{q}}\overline{H}^{q}+\frac{|\mathcal{E}(q)|\beta}{M}\sum_{n=1}^{N}{J^{qn}}^{T}J^{qn}\right)^{-1}V
\end{equation}
where 
$$
V=\frac{1}{M}\sum_{n=1}^{N}{J^{qn}}^{\text{T}}(A^{q}-\sum_{e\in\mathcal{E}(q)}C_{e}^{q}y_{e}^{qn(k)}+\beta\sum_{e\in\mathcal{E}(q)} C_{e}^{q}z_{e}^{qn(k)}).
$$
When $M\rightarrow \infty$, we have $\overline{H}^{q}=\frac{1}{M}\sum_{n=1}^{N}{J^{qn}}^{T}J^{qn}$ because
\begin{equation}
\begin{split}
&\nabla_{\theta^{i}}^2 D_{KL}(\pi^{i}_{old}(\cdot|o^{i};\theta^{i}_{old})||\pi^{i}(\cdot|o^{i};\theta^{i}))\left.\right|_{\theta^{i}=\theta^{i}_{old}} \\
=\ &\sum_{n=1}^{N}\nabla_{\theta^{i}}^2 D_{KL}(\pi^{i}_{old}(a^{n}|o^{i};\theta^{i}_{old})||\pi^{i}(a^{n}|o^{i};\theta^{i}))\left.\right|_{\theta^{i}=\theta^{i}_{old}} \\
=&\sum_{n=1}^{N}\mathbb{E}_{a\sim\pi^{i}_{old}}\left[\nabla_{\theta^{i}}\log(\Upsilon^{n}(\theta^{i})\nabla_{\theta^{i}}\log(\Upsilon^{n}(\theta^{i})\right]\left.\right|_{\theta^{i}=\theta^{i}_{old}}.
\end{split}
\end{equation}
Therefore, the optimal solution ${x^{q}}^{*}$ can be simplified as
\begin{equation}
\label{x_optimal}
{x^{q}}^{*} = \frac{1}{{\lambda^{q}}+|\mathcal{E}(q)|\beta} {\overline{H}^{q}}^{-1}V
\end{equation}
Substituting (\ref{x_optimal}) into (\ref{lagrange_x}), we derive
\begin{equation}
\begin{split}
p^{*}=&\max_{\lambda^{q}>0}\ f({x^{q}}^{*}, \lambda^{q})\\
=&\max_{\lambda^{q}>0}\ -\frac{1}{M({\lambda^{q}}+|\mathcal{E}(q)|\beta)}\sum_{n=1}^{N}{A^{q}}^{T} J^{qn}{\overline{H}^{q}}^{-1}V \\
+\frac{1}{M}&\sum_{e\in\mathcal{E}(q)}\sum_{n=1}^{N}{y_{e}^{qn}}^{T}\left(\frac{1}{{\lambda^{q}}+|\mathcal{E}(q)|\beta} C_{e}^{q}J^{qn}{\overline{H}^{q}}^{-1}V-z_{e}^{qn}\right) \\
+\frac{1}{M}&\sum_{e\in\mathcal{E}(q)}\sum_{n=1}^{N}\frac{\beta}{2}\parallel \frac{1}{{\lambda^{q}}+|\mathcal{E}(q)|\beta} C_{e}^{q}J^{qn}{\overline{H}^{q}}^{-1}V-z_{e}^{qn} \parallel_{2}^{2} \\
+\lambda^{q}&\left(\frac{1}{2(\lambda^{q}+|\mathcal{E}(q)|\beta)^2}V^{\text{T}}{\overline{H}^{q}}^{-1}V - \delta/N\right)
\end{split}
\end{equation}
Based on the first order condition, the optimal $\lambda^{q}$ should satisfy
\begin{equation}
\label{first_ordder_cond_lambda}
\begin{split}
&\frac{\partial f({x^{q}}^{*}, \lambda^{q})}{\partial\lambda^{q}} \\
=&\frac{1}{M({\lambda^{q}}+|\mathcal{E}(q)|\beta)^2}\sum_{n=1}^{N}{A^{q}}^{T} J^{qn}{\overline{H}^{q}}^{-1}V - \\
&\frac{1}{M({\lambda^{q}}+|\mathcal{E}(q)|\beta)^2}\sum_{e\in\mathcal{E}(q)}\sum_{n=1}^{N}C_{e}^{q}{y_{e}^{qn}}^{T}J^{qn}{\overline{H}^{q}}^{-1}V - \\
& \frac{n(\mathcal{E}(q))\beta}{(\lambda^{q}+|\mathcal{E}(q)|\beta)^3}V^{\text{T}}{\overline{H}^{q}}^{-1}V + (\text{Use } \overline{H}^{q}=\frac{1}{M}\sum_{n=1}^{N}{J^{qn}}^{T}J^{qn}) \\
& \frac{\beta}{M(\lambda^{q}+|\mathcal{E}(q)|\beta)^2}\sum_{e\in\mathcal{E}(q)}\sum_{n=1}^{N}C_{e}^{q}V^{T}{\overline{H}^{q}}^{-1}{J^{qn}}^{T}z_{e}^{qn} + \\
& \frac{|\mathcal{E}(q)|\beta-\lambda^{q}}{2(\lambda^{q}+|\mathcal{E}(q)|\beta)^3}V^{\text{T}}{\overline{H}^{q}}^{-1}V - \delta/N \\
=& \left(\frac{1}{(\lambda^{q}+|\mathcal{E}(q)|\beta)^2} -  \frac{1}{2(\lambda^{q}+|\mathcal{E}(q)|\beta)^2}\right)V^{\text{T}}{\overline{H}^{q}}^{-1}V - \delta/N \\
&(\text{Use}\ a^{T}{\overline{H}^{q}}^{-1}b = b^{T}{\overline{H}^{q}}^{-1}a \text{ since ${\overline{H}^{q}}^{-1}$ is symmetric}) \\
=& \frac{1}{2(\lambda^{q}+|\mathcal{E}(q)|\beta)^2}V^{\text{T}}{\overline{H}^{q}}^{-1}V - \delta/N \\
=&\ 0
\end{split}
\end{equation}
Solving Eq. (\ref{first_ordder_cond_lambda}), we get
\begin{equation}
\label{dual_optimal}
{\lambda^{q}}^{*} = \sqrt{\frac{V^{T}{\overline{H}^{q}}^{\ -1}V}{2\delta/N}}-|\mathcal{E}(q)|\beta.
\end{equation}
Substituting (\ref{dual_optimal}) into (\ref{x_optimal}), the optimal solution ${x^{q}}^{*}$ is 
\begin{equation}
{x^{q}}^{*} = \sqrt{\frac{2\delta/N}{V^{T}{\overline{H}^{q}}^{\ -1}V}}{\overline{H}^{q}}^{-1}V.
\end{equation}

Since $x^{q}=\theta^{q}-\theta^{q}_{old}$, the optimal update $\theta^{q(k+1)} :=\argmin_{\theta^{q}\in\Theta^{q}}\mathcal{L}_{\beta}(\bm{\theta},\bm{z}^{(k)},\bm{y}^{(k)})$ satisfies
\begin{equation}
\theta^{q(k+1)} = \theta^{q}_{old} + \sqrt{\frac{2\delta/N}{V^{T}{\overline{H}^{q}}^{\ -1}V}}{\overline{H}^{q}}^{-1}V.
\end{equation}
where
\begin{equation}
\notag
V=\frac{1}{M}\sum_{n=1}^{N}{J^{qn}}^{\text{T}}(A^{q}-\sum_{e\in\mathcal{E}(q)}C_{e}^{q}y_{e}^{qn(k)}+\beta\sum_{e\in\mathcal{E}(q)} C_{e}^{q}z_{e}^{qn(k)}).
\end{equation}

(2) Next, we consider the update of $z_{e}^{qn}$:
\begin{equation}
\min_{z_{e}^{qn}\in Z_{e}}\mathcal{L}_{\beta}(\bm{x},\bm{z},\bm{y}),\ q\in\mathcal{N}(e).
\end{equation}
where $Z_{e}^{n}$ $=\{z_{e}^{in},z_{e}^{jn}\vert z_{e}^{in}+z_{e}^{jn}=0, q=(i,j)\in\mathcal{N}(e)\}$.
This is a convex optimization problem. The optimal value $p^{*}$ satisfies
\begin{equation}
\label{lagrange}
p^{*}=\min_{z_{e}^{in},z_{e}^{jn}}\max_{\nu_{e}^{n}>0}\ \mathcal{L}_{\beta}(\bm{x},\bm{z},\bm{y})+{\nu_{e}^{n}}^{T}(z_{e}^{in}+z_{e}^{jn}).
\end{equation}
Based on the first order optimality conditions, the optimal $z_{e}^{in},z_{e}^{jn}$ should satisfy:
\begin{equation}
\label{zi}
-\frac{1}{M}y_{e}^{in} - \frac{1}{M}\beta\left(C_{e}^{i}J^{in}x^{i}-z_{e}^{in}\right) + \nu_{e}^{n} = 0
\end{equation}
\begin{equation}
\label{zj}
-\frac{1}{M}y_{e}^{jn} - \frac{1}{M}\beta\left(C_{e}^{j}J^{jn}x^{j}-z_{e}^{jn}\right) + \nu_{e}^{n} = 0
\end{equation}
Solving Eqs. (\ref{zi}) and (\ref{zj}), we get
\begin{equation}
{z_{e}^{in}}^{*}=\frac{1}{\beta}(y_{e}^{in}-M\nu_{e}^{n})+C_{e}^{i}J^{in}x^{i},
\end{equation}
\begin{equation}
{z_{e}^{jn}}^{*}=\frac{1}{\beta}(y_{e}^{jn}-M\nu_{e}^{n})+C_{e}^{j}J^{jn}x^{j}.
\end{equation}
Since ${z_{e}^{in}}^{*},{z_{e}^{jn}}^{*}$ should satisfy ${z_{e}^{in}}^{*}+{z_{e}^{jn}}^{*}=0$, we have 
\begin{equation}
\begin{split}
\frac{1}{\beta}(y_{e}^{in}+y_{e}^{jn})-\frac{2M}{\beta}\nu_{e}^{n}+(C_{e}^{i}J^{in}x^{i}+C_{e}^{j}J^{jn}x^{j})=0.
\end{split}
\end{equation}
Therefore, the optimal dual variable ${\nu_{e}^{n}}^{*}$ is\\
\begin{equation}
\begin{split}
{\nu_{e}^{n}}^{*}=&\frac{1}{2M}(y_{e}^{in}+y_{e}^{jn})+\frac{\beta}{2M}(C_{e}^{i}J^{in}x^{i}+C_{e}^{j}J^{jn}x^{j}) \\
=&\frac{1}{2M}\sum_{q\in\mathcal{N}(e)}\left[y_{e}^{qn}+\beta C_{e}^{q}J^{qn}x^{q}\right].
\end{split}
\end{equation}
Since $x^{q}=\theta^{q}-\theta^{q}_{old}$, the optimal update $z^{qn(k+1)}_{e} :=\argmin_{z^{qn}_{e}\in Z_{e}}\mathcal{L}_{\beta}(\bm{\theta}^{(k+1)},\bm{z},\bm{y}^{(k)})$ satisfies
\begin{equation}
z_{e}^{qn(k+1)}:=\ \frac{1}{\beta}(y_{e}^{qn(k)}-\nu_{e}^{n})+C_{e}^{q}J^{qn}(\theta^{q(k+1)}-\theta^{q}_{old}),
\end{equation}
where
\begin{equation}
\nu_{e}^{n}=\frac{1}{2}\sum_{q\in\mathcal{N}(e)}\left[y_{e}^{qn(k)}+\beta C_{e}^{q}J^{qn}(\theta^{q(k+1)}-\theta^{q}_{old})\right].
\end{equation}

(3) Substituting $z_{e}^{qn(k+1)}$ into the update
\begin{equation}
y^{qn(k+1)}_{e}:=y^{qn(k)}_{e}+\beta(C_{e}^{q}J^{qn}(\theta^{q(k+1)}-\theta^{q}_{old})-z_{e}^{qn(k+1)}),
\end{equation}
we get $y^{qn(k+1)}_{e}=\nu_{e}^{n}$.
\end{proof}


\ifCLASSOPTIONcaptionsoff
  \newpage
\fi



%

\bibliographystyle{IEEEtran}
\bibliography{ref}

\end{document}